\documentclass{article} % For LaTeX2e
\usepackage[final]{colm2026_conference}

\usepackage{microtype}
\usepackage{url}
\usepackage{fontawesome5}   % \faGithub

\usepackage{lineno}
\usepackage{latexsym}
\usepackage{multirow}
\usepackage{graphicx}
\usepackage{booktabs}
\usepackage{hyperref}       % hyperlinks
\usepackage{url}            % simple URL typesetting
\usepackage{amsfonts}       % blackboard math symbols
\usepackage{amsmath}        % text in math mode
\usepackage{amssymb}        % math symbols
\usepackage{nicefrac}       % compact symbols for 1/2, etc.
\usepackage{microtype}      % microtypography
\usepackage{xcolor}         % colors
\usepackage{algorithm}
\usepackage{algorithmic}
\usepackage{pifont}
\usepackage{subcaption}
\usepackage{graphicx}
\usepackage{tabularx}

\definecolor{darkblue}{rgb}{0, 0, 0.5}
\hypersetup{colorlinks=true, citecolor=darkblue, linkcolor=darkblue, urlcolor=darkblue}

\usepackage[T1]{fontenc}
\usepackage[utf8]{inputenc}

\newcommand{\officejs}[1]{OfficeJS}
\newcommand{\wtmbench}{\textsc{WTM-Bench}}
\newcommand{\wtmcorpus}{\textsc{WTM-Corpus}}

\title{$\circlearrowleft$ Back to the Future: A \emph{workbook time machine} for spreadsheet creation benchmarks}

\newcommand{\Author}[1]{\begin{tabular}[t]{c}\textbf{#1}\end{tabular}}

\author{
\makebox[\textwidth][c]{%
  \Author{Mansi Uniyal}\hspace{1.6em}
  \Author{Agamdeep Singh}\hspace{1.6em}
  \Author{Ananya Singha}\hspace{1.6em}
  \Author{Priyanshu Gupta}}
\AND
\makebox[\textwidth][c]{%
  \Author{Mukul Singh}\hspace{1.6em}
  \Author{Gust Verbruggen}\hspace{1.6em}
  \Author{Vu Le}\hspace{1.6em}
  \Author{Sumit Gulwani}}
\AND
\normalfont
\makebox[\textwidth][c]{%
  \begin{tabular}[t]{c}
    Microsoft\thanks{Email in order:
      \texttt{\{mansiuniyal, t-agasingh, ananyasingha, priyansgupta,
      singhmukul, gverbruggen, levu, sumitg\}@microsoft.com}}
  \end{tabular}}
}

\newtheorem{example}{Example}

\begin{document}

\ifcolmsubmission
\linenumbers
\fi

\maketitle

\begin{abstract}

We introduce the \emph{workbook time machine}, a pipeline that automatically creates benchmarks evaluating the ability of language models to create derived objects in spreadsheets (formulas, charts, pivot tables, and conditional formatting).
Applied to public workbook corpora, it produces \wtmcorpus{}--a collection of (input workbook, output workbook, query) triples spanning four artifact types and varying complexity.
From this corpus we curate \wtmbench{}, a 150-task evaluation benchmark with queries at three levels of specificity.
We evaluate existing spreadsheet manipulation agents and baselines on \wtmbench{} across artifact types, step complexity, and instruction granularity.
Our evaluations show that query specificity, agent orchestration, and interface API used to control spreadsheets play a big role in LLM performance on Excel tasks.
% \footnote{\href{https://figshare.com/s/2fa3ae21003453843e7c}{Link}: benchmark, generation and evaluation code}
\end{abstract}

\vspace{0.5em}
\begin{center}
\small
\faGithub~\href{https://aka.ms/wtm-bench}{\texttt{https://aka.ms/wtm-bench}}\\[0.25em]
\raisebox{-0.2em}{\includegraphics[height=1em]{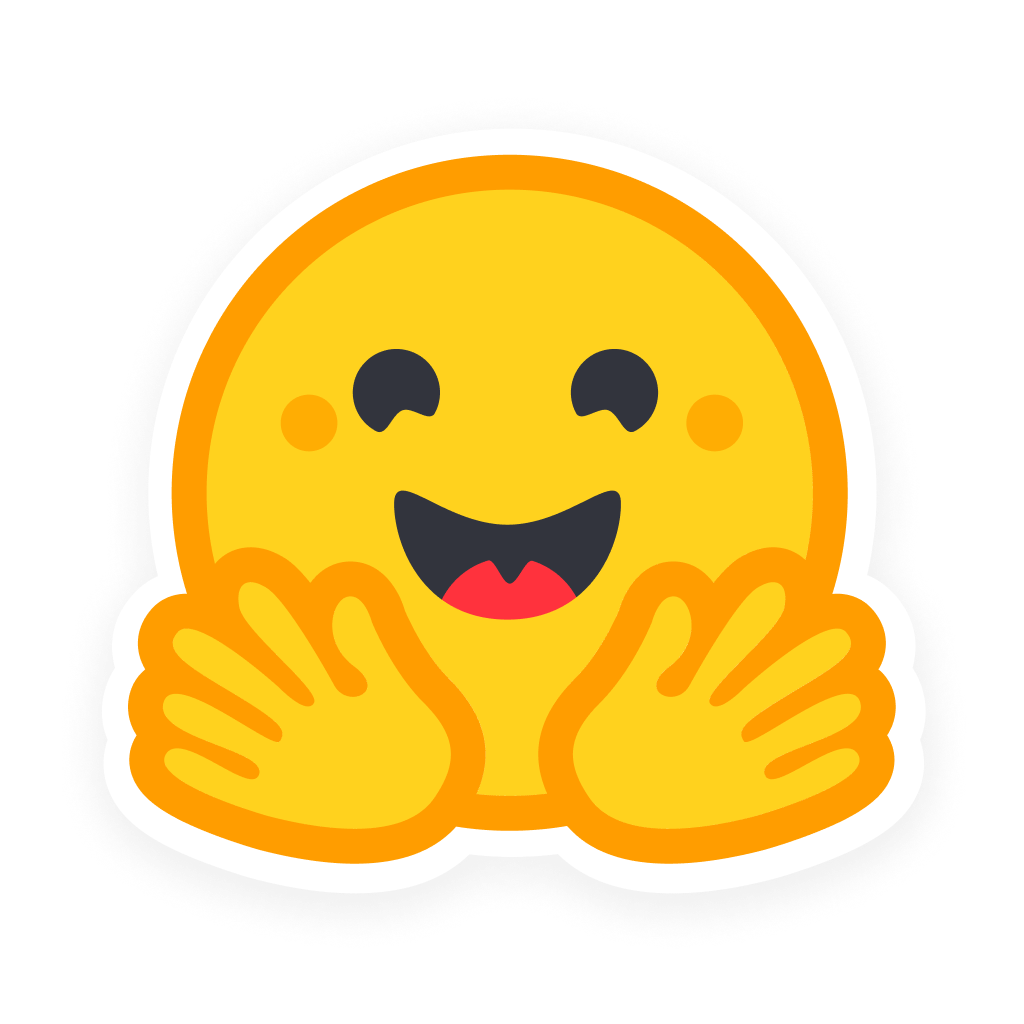}}~%
\href{https://aka.ms/hf-wtm-bench}{\texttt{https://aka.ms/hf-wtm-bench}}
\end{center}
\vspace{0.5em}

\section{Introduction}
\begin{figure}[htb]
    \centering
    \includegraphics[width=\linewidth]{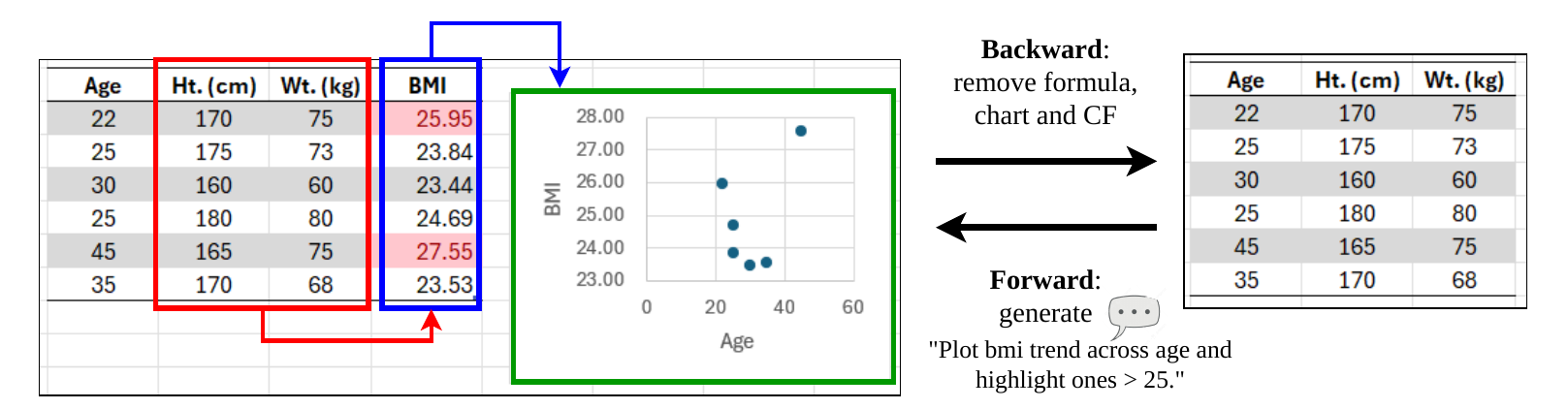}
    \caption{Example of creating a benchmark with the \emph{workbook time machine}, which removes objects from a workbook (left) and then generates an instruction that describes the removed objects.}
    \label{fig:example}
\end{figure}

Spreadsheets are the world's most widely used low-code platform, with over 750 million users~\citep{bendre2019faster, microsoft2024fy24q4}, serving as the primary computational tool for analysts, accountants, and domain experts who are not professional developers~\citep{hermans2016spreadsheets}.
Users routinely go beyond data entry, building formulas, charts, pivot tables, and conditional formatting rules--among other derived artifacts--that transform raw entries into structured analyses.

The recent success of coding agents in boosting developer productivity~\citep{peng2023copilot} suggests that similar gains are within reach for spreadsheet users, if models can reliably create the full range of \emph{derived artifacts} that real workbooks contain.
This has motivated enterprise agents for spreadsheet environments~\citep{microsoft2024copilot,anthropic2025claudeexcel,openai2025chatgptexcel}.
Yet progress has been hard to measure: existing benchmarks each cover a different slice of the problem--some use realistic workbooks but restrict tasks to formulas and data entry~\citep{spreadsheetbench}; others support diverse artifact types but operate on simple, hand-crafted files~\citep{sheetcopilot}; and still others target table reasoning rather than artifact creation~\citep{dong2024spreadsheetllm} or focus on conversational data analysis~\citep{condabench} (see Section~\ref{sec:related} for a detailed comparison).
To the best of our knowledge, no existing benchmark jointly evaluates on (i)~realistic, user-authored workbooks with complex structure (multiple sheets, non-standard layouts, cross-sheet dependencies); (ii)~multi-step creation of derived artifacts--formulas, charts, pivot tables, and conditional formatting; and (iii)~instructions at controllable levels of specificity.

We address this gap by \emph{reverse-engineering} real user work.
Taking inspiration from reverse curriculum generation approaches in reinforcement learning~\citep{florensa2017reverse,andrychowicz2017her}, which construct training distributions by working backward from goal states, we start from finished, user-authored spreadsheets and automatically reconstruct candidate edit histories--the orderings in which derived artifacts could plausibly have been created.
From these histories, we generate natural language instructions at multiple specificity levels for the same transformation, enabling systematic evaluation of how agents handle varying instructional detail.
We call this process the \emph{workbook time machine} (Figure~\ref{fig:example}).

\begin{example}\label{ex:introduction1}
Consider a spreadsheet with formulas computing BMI from height and weight columns, a conditional formatting rule highlighting high values, and a scatter chart plotting BMI against age.
The backward step strips these artifacts to recover the raw data.
The forward step generates instructions such as ``Calculate BMI and plot it against age'' (abstract) or ``In cell D2, enter \texttt{=10000*C2/(B2*B2)}, drag to D7, then create an XY scatter chart from A2:A7 vs D2:D7'' (fully specified).
\end{example}

This process yields three dimensions of controlled variation: (1)~\emph{artifact type}--which derived object must be created; (2)~\emph{step complexity}--how many intermediate artifacts the transformation requires; and (3)~\emph{instruction specificity}--how much detail the query provides.
Applied to the Enron~\citep{hermans2015enron} and FUSE~\citep{barik2015fuse} corpora, the pipeline produces \wtmcorpus{}: 8,931 queries over 2,977 unique tasks covering the major categories of Excel derived artifacts.
From this we curate \wtmbench{}, a balanced 150-task evaluation subset with near-uniform artifact distribution across three specificity levels.

We make the following contributions:
\begin{itemize}
    \item We introduce the \emph{workbook time machine}, a pipeline that reverse-engineers real user-authored spreadsheets into benchmark triples (input workbook, output workbook, query) by modeling candidate edit histories through dependency-aware DAG construction. Applying it to public corpora yields \wtmcorpus{}.
    \item From \wtmcorpus{}, we curate \wtmbench{}, a 150-task evaluation benchmark with controlled variation across artifact types, complexity, \& query specificity.
    \item We evaluate multiple agentic configurations across 6 \textit{frontier} models on \wtmbench{}, finding that (a)~API choice fundamentally shapes performance--OfficeJS provides richer Excel feature coverage while OpenPyXL struggles with charts and pivot tables; (b)~instruction specificity affects models asymmetrically--direct code generation excels with detailed instructions while agentic approaches better handle abstract queries; and (c)~artifact difficulty varies sharply--formulas are most tractable while pivot tables remain nearly unsolved.
\end{itemize}
\section{Related Works}
\label{sec:related}

\begin{table}
    \small
    \centering
    \caption{Comparison of benchmark properties
             $^*$ The paper mentions, yes, it is not found in practice in the dataset.}
    \label{tab:benchmark_comparison}
    \begin{tabular}{l@{}cccccc}
        \toprule
         && \textbf{SheetCopilotB.}& \textbf{InstructExcel}& \textbf{SheetRM}& \textbf{SpreadsheetB.}& \textbf{WTM}\\ \midrule
         \multirow{4}{*}{\textbf{Workbook }\hspace{2mm}}&real?& \ding{51} & \ding{51} & \ding{56} & \ding{51}  & \ding{51}\\
         &+ sheets?& \ding{51} & \ding{51} & \ding{56} & \ding{51}  &\ding{51} \\
         &+ tables?& \ding{56} & \ding{51} & \ding{56} & \ding{51}  &\ding{51} \\
         &+ info?& \ding{56} & \ding{51} & \ding{56} & \ding{56}  & \ding{51}  \\ \midrule
  \multirow{6}{*}{\textbf{Artifacts}}&formulas& \ding{51}
& \ding{51}
& \ding{51}
& \ding{51}
&\ding{51}
\\
  &charts& 
\ding{51}
& 
\ding{51}
& 
\ding{51}
& 
\ding{56}$^*$&\ding{51}  
\\
  &PT& 
\ding{51}  
& \ding{51}
& \ding{56}$^*$& \ding{56}$^*$&
\ding{51}  
\\
  &CF& 
\ding{51}  
& 
\ding{51}
& 
\ding{51}
& 
\ding{51}
&\ding{51}  
\\
  &cell manip.& 
\ding{51}  
& \ding{51}
& \ding{51}
& \ding{51}
&
\ding{56}  \\
  &comb.& 
\ding{51}& \ding{56} & \ding{51}
& \ding{51}&\ding{51}  
\\
        \bottomrule
    \end{tabular}
\end{table}

\paragraph{Spreadsheet benchmarks.}
Existing spreadsheet benchmarks each cover a different slice of the problem space (Table~\ref{tab:benchmark_comparison}).
SheetCopilotBench~\citep{sheetcopilot} supports diverse artifact types (charts, pivot tables, conditional formatting) but operates on hand-crafted workbooks that lack the structural complexity--multiple sheets, non-standard layouts, cross-sheet references--of real user files.
InstructExcel~\citep{payan2023instructexcel} scales to thousands of instruction-code pairs and uses real workbooks, yet each task targets a single isolated operation; multi-step workflows and varying instruction granularity are not supported.
SpreadsheetBench~\citep{spreadsheetbench} grounds evaluation in realistic, user-authored workbooks sourced from forums, but its task scope is limited to data entry and formula manipulation--charts and pivot tables are absent in practice despite being mentioned.
SheetRM~\citep{sheetagent} introduces a reward model for spreadsheet agents trained on synthetic workbooks, which limits transferability to the messy, multi-table environments encountered in practice.
ConDABench~\citep{condabench} evaluates LLMs on \emph{conversational} data analysis tasks requiring multi-turn interaction and disambiguation of under-specified goals, but it targets analytical insights over tabular data rather than the creation of spreadsheet artifacts (formulas, charts, pivot tables) that \wtmbench{} focuses on and avoids low-value operation like manual cell-manipulations.
In contrast, \wtmbench{} combines real workbooks with multi-artifact creation tasks and controllable instruction specificity.
\paragraph{Spreadsheet understanding and code generation.}
SpreadsheetLLM~\citep{dong2024spreadsheetllm} develops encoding schemes that preserve spatial relationships for spreadsheet question answering, while TableTalk~\citep{tabletalk} enables natural language interaction with structured tables--both are read-only and do not modify workbook content.
SheetMind~\citep{sheetmind} reasons over cell dependencies and formula relationships, an ability we leverage in our dependency-aware pruning (Section~\ref{sec:backward}).
On the code generation side, approaches for programmatic spreadsheet control~\citep{payan2023instructexcel,sheetmind} and multi-step task planning~\citep{sheetcopilot,sheetagent} address orchestration of complex workflows--but all assume clean starting states rather than the artifact-rich environments of real workbooks. More broadly, code-generating LLM agents~\citep{yang2024llmagents} have shown promise for automating multi-step tasks, yet spreadsheet-specific challenges (non-standard layouts, cross-sheet dependencies, API heterogeneity) remain underexplored.

\paragraph{Backward generation and self-improvement.}
Our reverse-engineering view is also related to methods that construct learning problems by working backward from known goal states. Reverse curriculum learning and hindsight experience replay~\citep{florensa2017reverse,andrychowicz2017her} use reachable goals to densify supervision, while agentic self-debugging and action-observation loops~\citep{chen2023selfdebug,yao2023react} use intermediate feedback to refine multi-step solutions. The Edit DAG differs in purpose: it is not a training-time search policy, but a data-construction mechanism that enumerates reachable spreadsheet transformations from real final workbooks.

\section{Problem Formulation}
\label{sec:problem}

Let $\mathcal{W} = \{W_1, W_2, \ldots, W_m\}$ be a corpus of Excel workbooks.
Each workbook $W \in \mathcal{W}$ consists of raw data $D$ and a set of \emph{derived artifacts} $C = \{c_1, \ldots, c_n\}$--formulas, charts, pivot tables, conditional formatting rules--built on top of~$D$.
We write $W = \{D\} \cup C$.

Given $\mathcal{W}$, our goal is to produce a benchmark
\[
\mathcal{B} = \bigl\{(W^{\text{in}}_k,\; W^{\text{out}}_k,\; q_k)\bigr\}_{k=1}^{K},
\]
where $W^{\text{in}}_k \subset W^{\text{out}}_k \subseteq W$ are intermediate workbook states, that differ by $\geq$ 1 artifacts, and $q_k$ is a natural language instructions describing the transformation from $W^{\text{in}}_k$ to $W^{\text{out}}_k$.
At evaluation time, a model receives $(W^{\text{in}}_k, q_k)$ and must produce a state matching $W^{\text{out}}_k$.

\section{Workbook Time Machine}
\label{sec:method}

Given a final workbook, our method operates in two passes.
In the \emph{backward step}, we decompose the workbook by stripping its derived artifacts and estimating the various \emph{candidate edit histories}--timelines of edits that could have produced the final workbook.
In the \emph{forward step}, we sample transformations from these candidate histories and generate natural language instructions at varying levels of specificity.

\begin{figure}[htb]
    \centering
    \includegraphics[width=\linewidth]{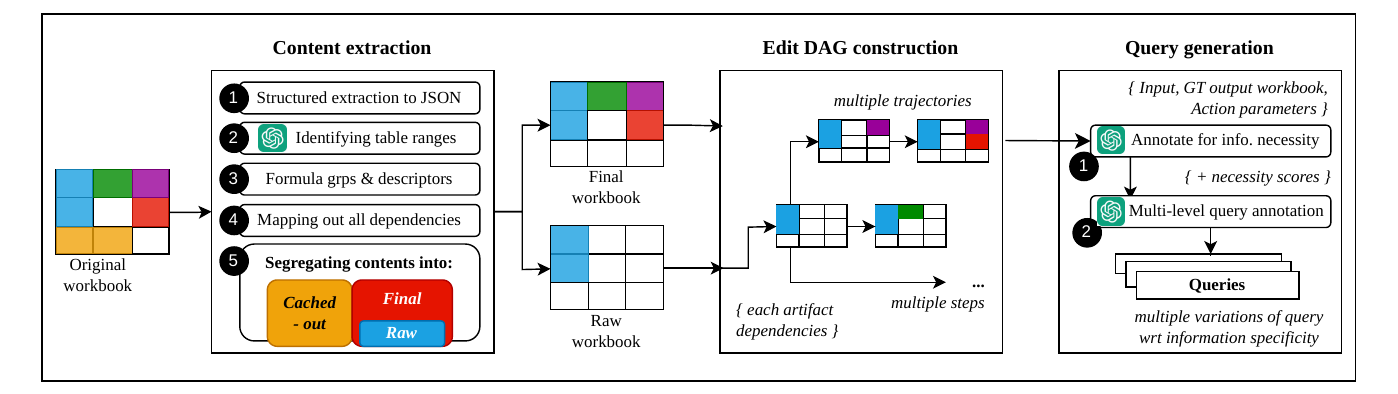}
    \caption{Generation pipeline flowchart}
    \label{fig:flowchart}
\vspace{-10pt}
\end{figure}

\subsection{Component extraction.}
Recall from Section~\ref{sec:problem} that a workbook $W = \{D\} \cup C$ comprises raw data $D$ and derived artifacts $C = \{c_1, \ldots, c_n\}$.
Starting from the final workbook $W$, the backward step strips away derived artifacts to recover the raw state $W^0 = \{D\}$, and then reconstructs the candidate edit histories--all semantically valid orderings in which the artifacts could have been added back.
We construct $W^0$ by stripping all derived artifacts from $W$.
Beyond straightforward removal, we apply two heuristics to ensure clean extraction.
First, we perform \emph{formula grouping}: spreadsheet features such as FlashFill~\citep{gulwani2011flashfill} and formula drag allow users to replicate a formula across contiguous ranges, so we anonymize cell references and group adjacent formulas that share the same template into a single \emph{formula group}, substantially reducing the number of artifacts.
Second, we perform \emph{semantic descriptor mapping}: label cells that describe an adjacent formula--for example, a ``Total'' cell next to \texttt{=SUM()}--leak information about the target transformation and must be associated with the source state and removed if the component is removed.
We use an LLM to identify semantic table ranges and mark row and column headers that serve as descriptors for formula groups.

\subsection{Backward Step: Decomposition}
\label{sec:backward}

\paragraph{Edit DAG construction.}
We represent these candidate histories compactly as a Directed Acyclic Graph (DAG).
Let the true edit history of the workbook be $\mathcal{T} = (W^0, W^1, W^2, \ldots, W)$.
Since $\mathcal{T}$ is not available, we model all possible candidate edit histories via the \emph{Edit DAG} $\mathcal{G} = (V, E)$, where $V$ comprises all workbook states obtainable by adding subsets of $\{c_1, \ldots, c_n\}$ to $W^0$, and a directed edge $(W^i, W^j) \in E$ exists iff $W^j = W^i \cup \{c_k\}$ for some artifact $c_k$.
Each path from $W^0$ to $W$ in $\mathcal{G}$ corresponds to one candidate edit history.

\paragraph{Dependency-aware pruning.}
Na\"ively, a workbook with $n$ artifacts admits up to $n!$ orderings.
However, not all orderings are semantically valid--a chart built on a derived column cannot precede the creation of that column.
Building on prior work on cell dependency analysis~\citep{formulagraphs,sheetmind}, we capture such constraints through a \emph{artifact dependency graph} $\mathcal{G}_D = (\{c_1, \ldots, c_n\}, E_D)$, where $(c_i, c_j) \in E_D$ when the input range of $c_j$ overlaps the output range of $c_i$.
We prune every edge $(W^i, W^i \cup \{c_j\}) \in E$ for which at least one prerequisite of $c_j$ is absent from $W^i$.
If the pruned graph is no longer fully connected, we retain the connected artifact containing $W$, preferentially selecting artifact-dense states.
We call the result the \emph{Dependency-Pruned DAG} $\mathcal{G}_P$.

\paragraph{Enumerating candidate histories.}
We enumerate all sub-paths in $\mathcal{G}_P$ as candidate edit histories.
Crucially, these sub-paths need not originate at $W^0$: a candidate history may begin at any intermediate state that already contains a subset of derived artifacts.
This allows the benchmark to include the realistic scenario in which a user continues building on a partially constructed workbook rather than starting from raw data.

\begin{table*}[htb]
\vspace{-7pt}
    \centering
    \small
    \caption{Different levels of instructions with increasing levels of abstractness.}
    \label{tab:instructions}
    \begin{tabularx}{\textwidth}{lX}
        \toprule
         1&In Sheet1, calculate the BMI for each individual by using the formula BMI = 10000 * Wt. (kg) / (Ht. (cm) * Ht. (cm)) and place the results in column D, starting from D2 to D7. Ensure that the header `BMI' is added in cell D1. Then, create a scatter chart with a line marker subtype, using `Age' from A2:A7 as the x-axis and the calculated `BMI' from D2:D7 as the y-axis. Title the chart `Age' and label the axes as `Age' and `BMI'.  \\
         2&Calculate BMI for each person in Sheet1 using their height and weight, and place the results in column D. Add a header `BMI' in D1. Create a scatter chart with `Age' as the x-axis and the calculated `BMI' as the y-axis, titled `Age'. \\
         3&Create a scatter chart in Sheet1 using Age and BMI data.  \\
         \bottomrule
    \end{tabularx}
\end{table*}

\subsection{Forward Step: Query Generation}
\label{sec:forward}

Given the candidate edit histories enumerated from $\mathcal{G}_P$, the forward step selects endpoint pairs as input-output tuples $(W^{\text{in}}, W^{\text{out}})$ and generates natural language instructions for each transformation.

\paragraph{Action extraction and necessity scoring.}
\vspace{-2pt}

For each transformation, we extract a structured \emph{action} representation consisting of key-value pairs that describe the relevant parameters (e.g., chart type, data range, axis labels, formula expression).
An LLM assigns a \emph{necessity score} $s \in \{0, 0.5, 1\}$ to each parameter, where $s{=}1$ denotes parameters essential for identifying the task and $s{=}0$ denotes incidental details.
This yields action representations at three levels of specificity: fully specified (all parameters), moderately specified ($s > 0$), and minimal ($s{=}1$).

\paragraph{Natural language instruction generation.}
\vspace{-2pt}

Each filtered action variant is provided as conditioning context to an LLM, which generates a corresponding natural language instruction.
This produces instructions ranging from fully specified to abstract, as illustrated in Table~\ref{tab:instructions}.

\section{Benchmark Construction}
\label{sec:dataset}

The workbook time machine generates benchmark instances from real-world Excel workbooks, producing 8{,}931 natural-language queries over 2{,}977 unique tasks from the Enron~\citep{hermans2015enron} and FUSE~\citep{barik2015fuse} corpora (\wtmcorpus{}). However, this distribution suffers from severe class imbalance (67.5\% formulas vs. 0.8\% pivot tables), making LLM evaluation both computationally prohibitive and methodologically problematic. We therefore curate \wtmbench{}, a 150-task evaluation subset (450 queries across three specificity levels) via principled lexicographic sampling that achieves near-uniform task-type distribution while preserving semantic properties.

\begin{figure*}[htbp]
\vspace{-7pt}
    \centering
    \includegraphics[width=\linewidth]{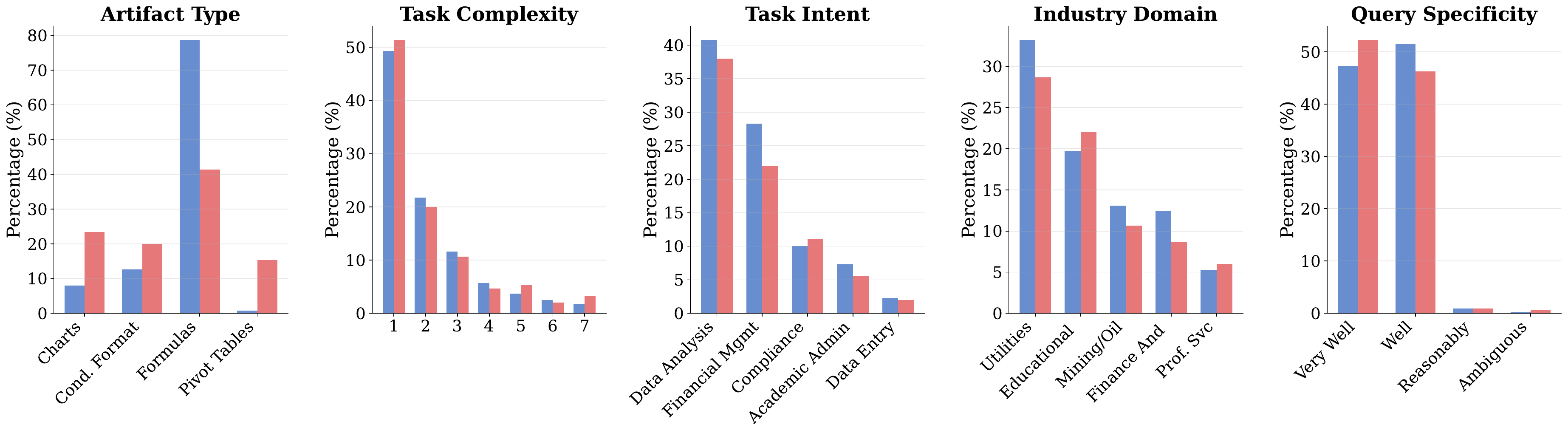}
    \caption{\wtmbench{} achieves better artifact distribution while preserving other attributes. 
             \textcolor{blue}{Blue} bars for \wtmcorpus{} and \textcolor{red}{red} bars for \wtmbench{} dataset.}
    \label{fig:eval_construction}
\vspace{-10pt}
\end{figure*}

\paragraph{Addressing distribution imbalance.}
Our primary design challenge was the extreme skew in the original distribution where formulas dominate (67.5\%) while pivot tables represent only 0.8\% of tasks. This imbalance would render aggregate performance metrics meaningless, as they would primarily reflect formula-writing capability rather than comprehensive Excel automation skills. Our lexicographic sampling strategy achieves dramatic rebalancing: 41.3\% formulas, 23.3\% charts, 20.0\% conditional formatting, and 15.3\% pivot tables. 

\paragraph{Multi-level query design rationale.}
\vspace{-2pt}

We hypothesized that sequential generation (Level 1→2→3) sequentially in 1 LLM call, would provide superior information retention compared to direct Level 3 generation. This design choice was validated through systematic comparison on 75 randomly sampled tasks, confirming that the multi-level approach achieves better information preservation with reduced data leakage (detailed analysis in Appendix~\ref{app:design_validation}).

\vspace{-2pt}
\paragraph{Computational efficiency considerations.}
Pipeline construction scales predictably with workbook complexity: $\text{LLM}_{\text{total}} = \text{LLM}_{\text{extract}} + \text{LLM}_{\text{annotate}}$, where annotation cost grows as $2 \times \sum \text{\# permutations per trajectory}$. This ensures richer workbooks amortize annotation costs across proportionally more training examples. More details are in Appendix~\ref{app:construction_efficiency}.

\paragraph{Quality validation methodology.}
\vspace{-2pt}

We conducted systematic quality evaluation using 3 human annotators across 3 dimensions with inter-annotator reliability analysis to characterize query naturalness and completeness (details in Appendix~\ref{sec:quality}). Our sampling strategy maintains semantic equivalence across the complete dataset, with intent distribution remaining stable across query levels, validating that the 3 query variants target the same underlying transformations despite varying specificity.

\section{Benchmark Analysis: \wtmbench{}}
\label{sec:analysis}

\paragraph{Task distribution insights.}
\vspace{-2pt}

\begin{figure*}[htbp]
    \centering
    \includegraphics[width=\linewidth]{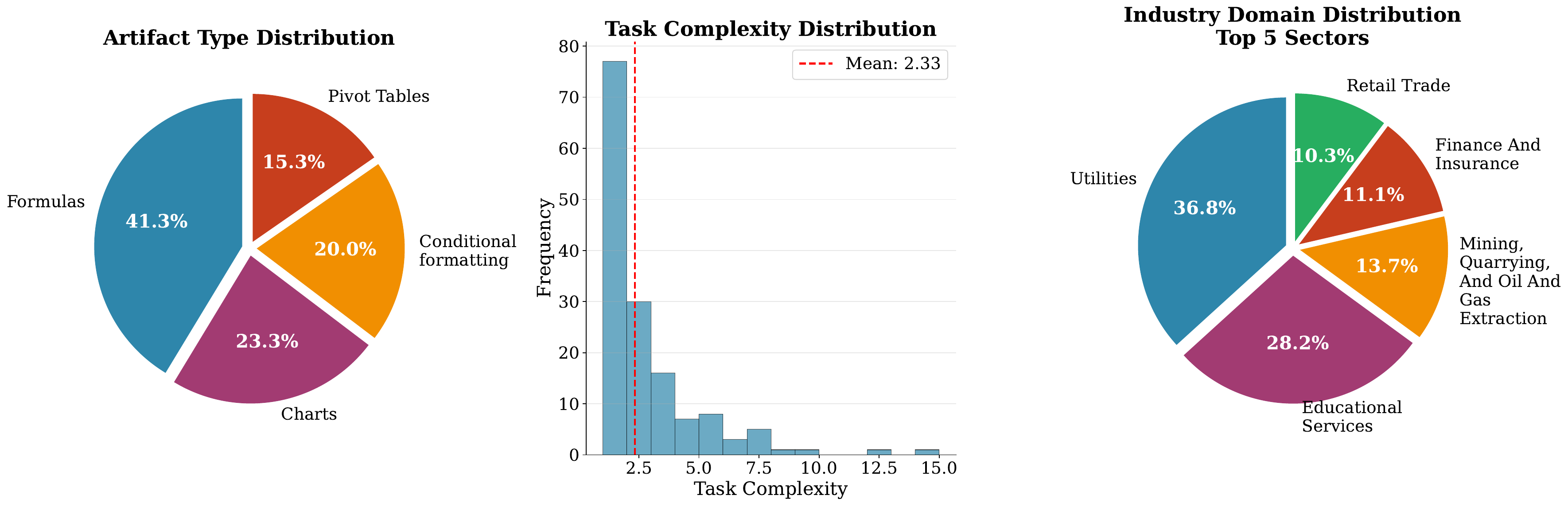}
    \caption{Task distribution analysis over \wtmbench{}.}
    \label{fig:task_overview}
\end{figure*}

Figure \ref{fig:task_overview} shows how our rebalanced benchmark provides diagnostic capability across Excel's functional spectrum rather than primarily testing formula proficiency. The complexity distribution reveals authentic real-world patterns: while most tasks (51.3\%) involve simple 1-2 artifact transformations, a substantial tail extends to 14-artifact workflows, enabling analysis of where reasoning capabilities break down as complexity increases.  
We define this expected number of artifact transformation as \emph{Task Complexity} of any data pair.

Diversity with respect to workbook selection is also captured in the industry domain. Where top-5 major sectors serve as a natural regularizer against domain-specific overfitting. This taxonomy analysis uses sector categorization from NAICS~\citep{naics2022} and uses an LLM for classification.

\paragraph{Multi-level design as capability probe.}
\vspace{-2pt}

\begin{figure*}[htbp]
    \centering
    \begin{subfigure}[b]{0.48\textwidth}
        \centering
        \includegraphics[width=\linewidth]{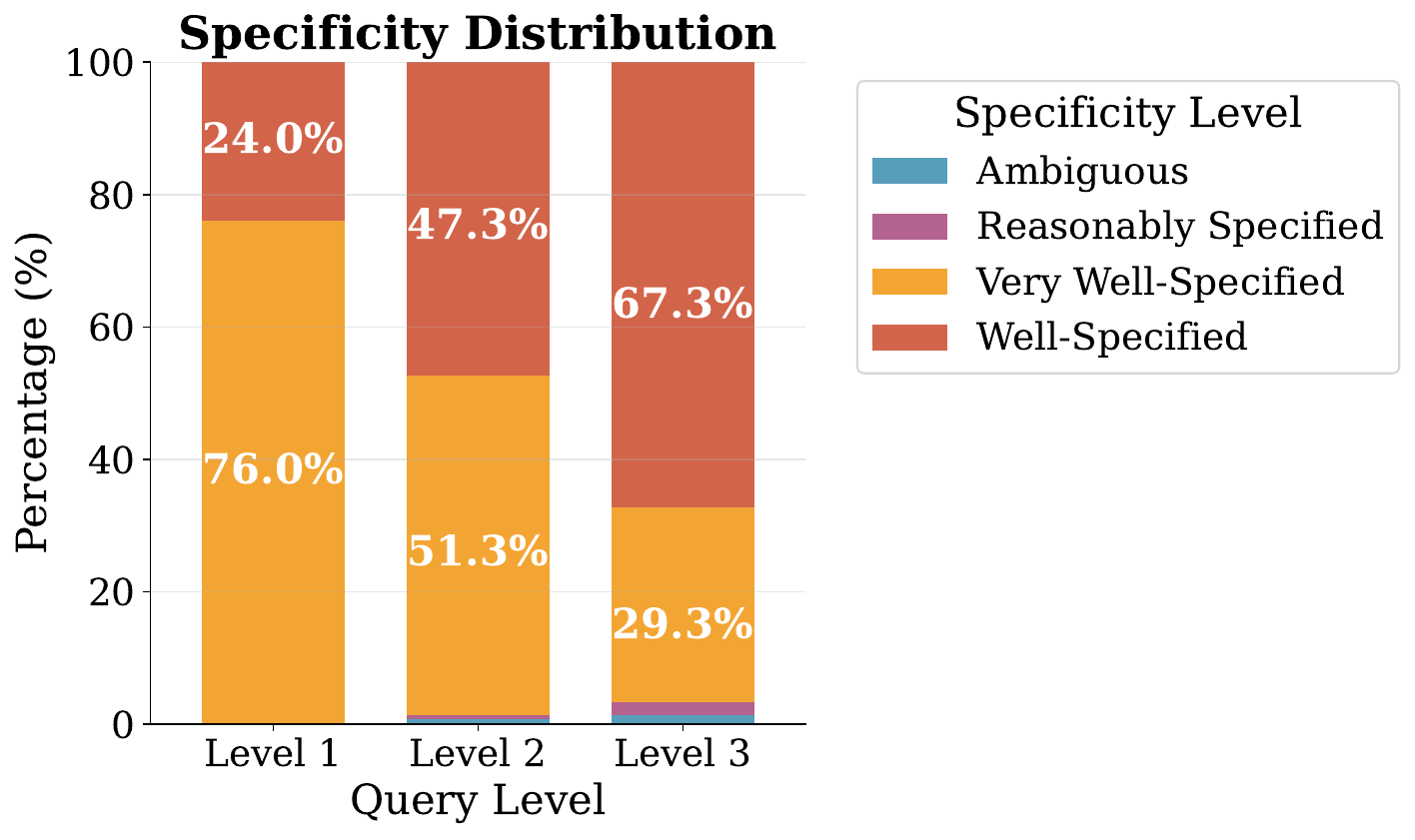}
        \caption{Specificity gradient across levels}
        \label{fig:specificity_stacked}
    \end{subfigure}
    \hfill
    \begin{subfigure}[b]{0.48\textwidth}
        \centering
        \includegraphics[width=\linewidth]{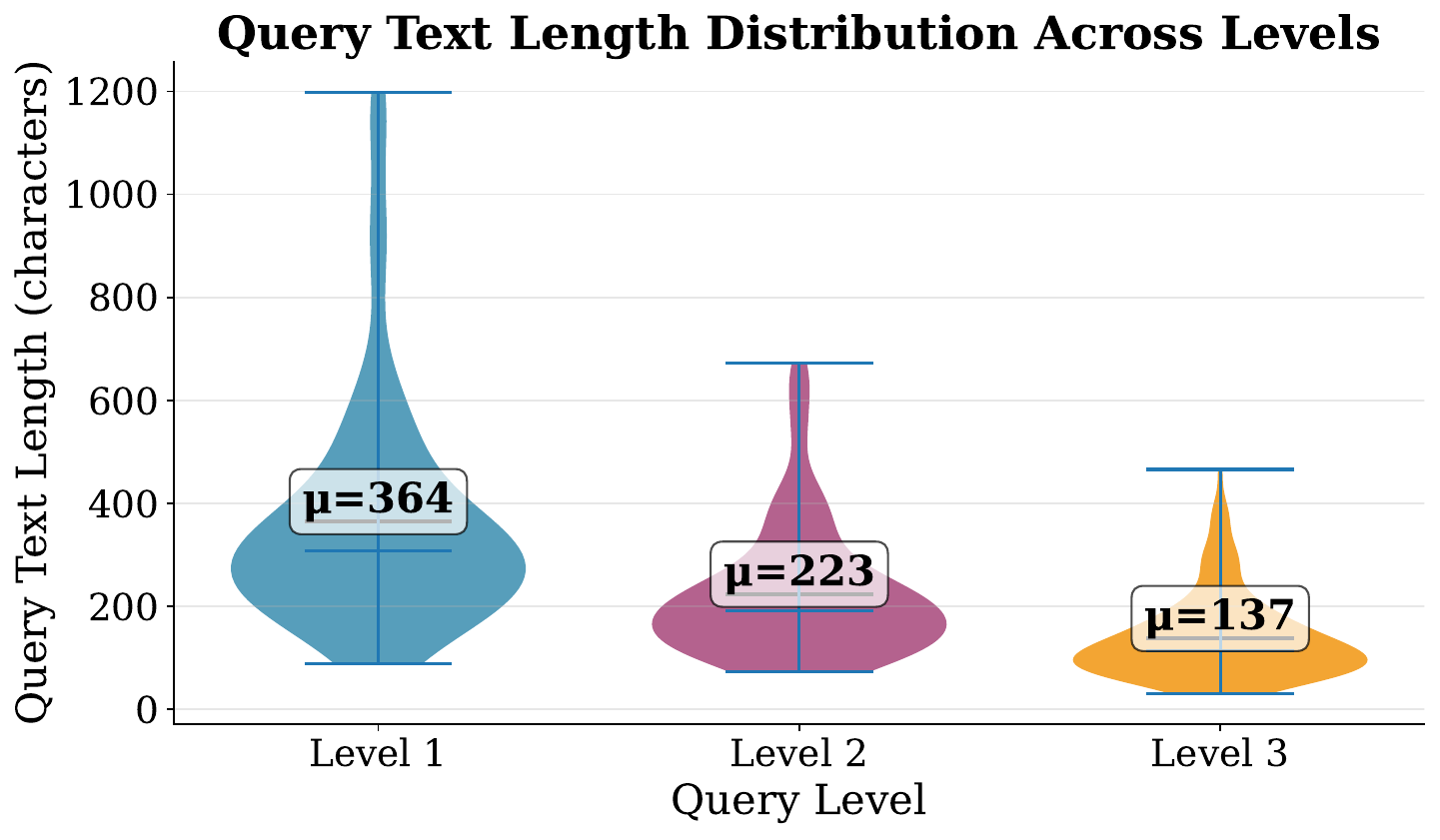}
        \caption{Query length distributions}
        \label{fig:query_length_violin}
    \end{subfigure}
    \caption{Query characteristic analysis.}
    \label{fig:query_design}
\vspace{-15pt}
\end{figure*}

Figure \ref{fig:query_design} plots the 3-level specificity gradient, creating a controlled experimental setting that isolates instruction-following from task execution capabilities. The systematic degradation of specificity from detailed (Level 1) to concise (Level 3) instructions reveals whether models fail due to task complexity or insufficient context. This diagnostic dimension also gets missed in single-level benchmarks. Thus enabling precise identification of model limitations: strong Level 1 performance with poor Level 3 results indicates gap-filling rather than reasoning deficits.

The annotation study suggests a trade-off between brevity and completeness: shorter queries tend to receive higher human-likeness scores and lower completeness scores, although these dimensions exhibit only moderate agreement and should be interpreted as indicative rather than definitive. This tension helps explain why Level 3 queries pose particular challenges, as they reflect realistic human communication patterns that rely on workbook context and spreadsheet conventions.

\section{Experiment Setup}

\vspace{-2pt}

\paragraph{Models.} We evaluate 6 state-of-the-art language models spanning two major families: Anthropic (Claude Haiku 4.5, Claude Sonnet 4.5, Claude Opus 4.6) and OpenAI (GPT-4.1 Mini, GPT-5.2 Reasoning, GPT-5.4 Reasoning).

\paragraph{Orchestration.}
\vspace{-2pt}
We evaluate popular Python-based Excel LLM orchestration frameworks--Sheet Agent, SheetCopilot, and SpreadsheetBench--with GPT 5.4 Reasoning. 
However, motivated by advances in tool-execution agents \citep{jimenez2024swebench, merrill2026terminalbenchbenchmarkingagentshard}, we propose a custom orchestration framework adopting modern tool-calling best practices in which the model iteratively emits structured code calls, observes execution results, and self-debugs across up to 10 turns. We evaluate across three Excel interface APIs: Python (\texttt{openpyxl}), VBA, and Office.js. Full implementation details are provided in Appendix~\ref{app:exp_details}.

\paragraph{Sampling parameters.}
Generation uses GPT-5.4 Reasoning to author the natural-language query variants and necessity scores. During evaluation, GPT reasoning models use their default sampling configuration and Claude models use temperature 1.0, following Anthropic's benchmark setting. The turn cap is 10 for all multi-turn configurations.

\paragraph{Evaluation}
\vspace{-2pt}

We evaluate performance on \wtmbench{} using two complementary metrics. The \textbf{Soft Score} is a continuous reward in $[0, 1]$ measuring partial task completion, computed as a weighted combination of component-level similarity scores across cells, formulas, charts, pivot tables, and conditional formatting (see Appendix~\ref{app:metrics} for details). The \textbf{Hard Score} is the fraction of tasks where the agent achieves a perfect Soft Score of 1.0, capturing complete and exact task reconstruction. All grading is deterministic and programmatic: no LLM is used to judge model outputs, and the query generator is never consulted during evaluation. We present both metrics as \% scores (scaled 0--100). Additionally, \textbf{\# Turns} counts the code execution turns, with execution feedback, used by the model to complete a task.
\section{Results and Analysis}
\begin{table}[t]
\small
    \centering
    \setlength{\tabcolsep}{10pt}
    \renewcommand{\arraystretch}{1.15}
    \begin{tabular}{llccc}
        \toprule
        \textbf{Orchestration} & \textbf{Model} & \textbf{ \# Turns} & \textbf{Hard Score (\%)} & \textbf{Soft Score (\%)} \\
        \midrule
        \midrule
        Python          & ClaudeHaiku45         & 5.45 & 10.0                   & 33.7          \\
        (OpenPyXl)      & ClaudeSonnet45        & 5.02 & 16.7                   & 40.4          \\
                        & ClaudeOpus46          & 4.31 & \underline{18.7}       & \textbf{48.6} \\
                        & Gpt41Mini             & 2.27 & 7.3                    & 26.0          \\
                        & Gpt52Reasoning        & 3.74 & 15.3                   & 44.6          \\
                        & Gpt54Reasoning        & 3.33 & \textbf{20.0}          & \underline{46.3} \\
        \midrule
        OfficeJS        & ClaudeHaiku45         & 4.61 & 12.0                   & 36.7          \\
                        & ClaudeSonnet45        & 4.18 & \textbf{22.0}          & \underline{46.1} \\
                        & ClaudeOpus46          & 4.33 & \underline{19.3}       & \textbf{48.1} \\
                        & Gpt41Mini             & 2.69 & 9.3                    & 29.3          \\
                        & Gpt52Reasoning        & 3.07 & 14.4                   & 44.3          \\
                        & Gpt54Reasoning        & 2.63 & 14.7                   & 40.9          \\
        \midrule
        VBA             & ClaudeHaiku45         & 3.16 & 13.0                   & 38.4          \\
                        & ClaudeSonnet45        & 2.92 & \underline{32.7}       & 51.3          \\
                        & ClaudeOpus46          & 3.50 & \textbf{33.7}          & \textbf{54.0} \\
                        & Gpt41Mini             & 2.20 & 12.6                   & 30.5          \\
                        & Gpt52Reasoning        & 3.09 & 30.9                   & \underline{53.9} \\
                        & Gpt54Reasoning        & 2.31 & 22.8                   & 39.0          \\
        \midrule
        SheetCopilot & Gpt54Reasoning     &5.03& 12.7  & 34.9 \\
        SheetAgent & Gpt54Reasoning  &9.15& 12.0  & 32.6 \\
        SpreadsheetBench & Gpt54Reasoning  & 1.0 & 3.3   & 8.52          \\
        \bottomrule
    \end{tabular}
    \caption{Results for popular frontier models and orchestration frameworks on level 3 queries. The best-performing scores are marked in \textbf{bold}, and the second best in \underline{underline}.}
    \label{tab:l5_results}
\end{table}
%%%%%%%%%%%%%%%%%%%%%%%%%%%%%%%%%%%%%%%%%%%%%%%%%%%%%%%%%%%%
% NEW RESULTS SECTION — for comparison with the above
%%%%%%%%%%%%%%%%%%%%%%%%%%%%%%%%%%%%%%%%%%%%%%%%%%%%%%%%%%%%
\iftrue  % <<< toggle: change to \iftrue to render new version, \iffalse to hide
% \section{Results and Analysis (v2 — proposed rewrite)}
% \label{sec:results_v2}

We evaluate 6 frontier models of various sizes on \wtmbench{} using our tool-calling orchestration across three API tooling, and compare against three established agent frameworks. Our analysis reveals three key findings: \textbf{(1)}~agent design matters more than model scale---our orchestration with structured state observation outperforms existing frameworks by 7--17~pp in Hard Score using the same backbone model; \textbf{(2)}~all models fail catastrophically on pivot tables ($\leq$10\% Soft Score), exposing a fundamental reasoning gap; and \textbf{(3)}~performance degrades monotonically with query abstraction and task complexity, validating \wtmbench{} as a multi-dimensional diagnostic benchmark.

\paragraph{Agent Design and Orchestration}
\vspace{-2pt}
Table~\ref{tab:l5_results} reports results on Level~3 queries. Using GPT-5.4~Reasoning as a common backbone, our simple orchestration achieves 20.0\% Hard Score and 46.3\% Soft Score (Python API), outperforming existing specialized agents such as SheetCopilot (12.7\% Hard, 34.9\% Soft), SheetAgent (12.0\% Hard, 32.6\% Soft), and SpreadsheetBench (3.3\% Hard, 8.5\% Soft). These gaps highlight the importance of agent architecture over raw model capability.
Notably, SpreadsheetBench's single-turn, no-feedback approach yields the lowest performance. This highlights that models require structured state observation: our orchestration surfaces the post-execution workbook state at each turn, enabling the model to ground its next action in the actual spreadsheet rather than relying on just prior code generation.
\paragraph{Model and API Analysis}
\vspace{-2pt}
The best-performing configuration overall is Claude~Opus~4.6 with VBA (33.7\% Hard, 54.0\% Soft). Across APIs, we observe that VBA and OfficeJS consistently outperform Python for the same model, which we attribute to richer native Excel primitives.  For instance, creating a pivot table in VBA requires a single \texttt{PivotTable} call, whereas Python requires orchestrating Pandas for the pivot operation and OpenPyXL for writing the result---a multi-step workflow that compounds errors.

This effect is model-dependent: Claude models benefit substantially from richer APIs (Opus: +15.0~pp Hard Score from Python to VBA; Sonnet: +16.0~pp), whereas GPT-5.4~Reasoning shows only a modest gain (+2.8~pp). This suggests that Claude models are better at leveraging unfamiliar API surfaces, possibly due to stronger instruction-following capabilities.

Turn efficiency reveals an additional pattern: smaller models (GPT-4.1~Mini, Claude~Haiku) use fewer turns (2.2--3.2) but score lower, while flagship models use 3--5 turns more productively. Independent Python/OpenPyXL re-runs preserve model ordering, with Hard Score standard deviations at or below 3.5 percentage points (Appendix~\ref{app:robustness}), suggesting that the headline differences are not single-run artifacts.
\begin{figure*}[htbp]
    \centering
    \begin{subfigure}[b]{0.50\textwidth}
        \centering
        \includegraphics[width=\linewidth]{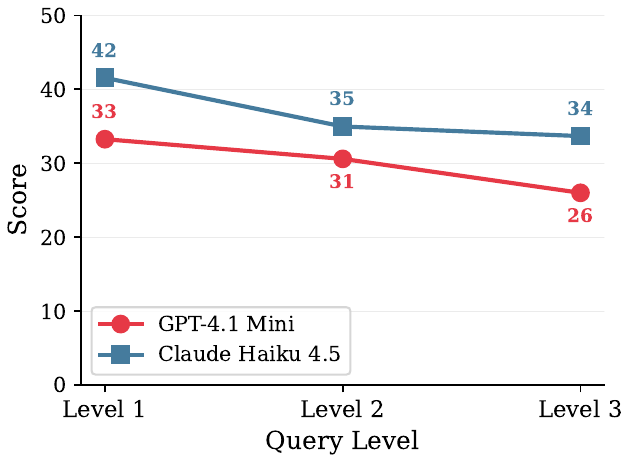}
        \caption{Soft Score for different query levels}
        \label{fig:level_score}
    \end{subfigure}
    \hfill
    \begin{subfigure}[b]{0.44\textwidth}
        \centering
        \includegraphics[width=\linewidth]{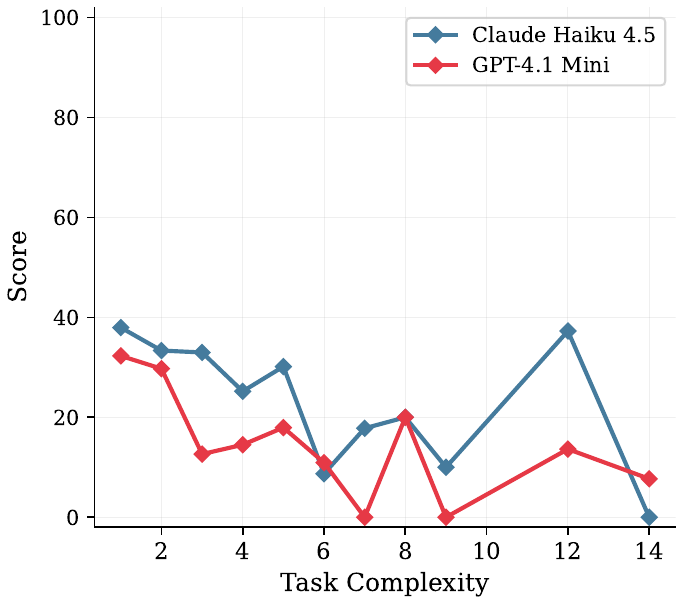}
        \caption{Soft Scores for different task complexities }
        \label{fig:steps_score}
    \end{subfigure}
    \caption{Soft scores decrease with complexity; Haiku 4.5 outperforming GPT-4.1 Mini.}
    \label{fig:plots_added}
\vspace{-10pt}
\end{figure*}

\paragraph{Effect of Query Specificity}
\vspace{-2pt}

Figure~\ref{fig:level_score} shows Soft Scores across the three query levels. Performance degrades monotonically from Level~1 (verbose, detailed) to Level~3 (concise, abstract). This aligns with our dataset analysis: the proportion of ``Very Well-Specified'' queries drops 4.4$\times$ from Level~1 to Level~3, and mean query length shrinks from 357 to 139 characters. 
The multi-level design also provides a natural curriculum for future fine-tuning---models can be trained on Level~1 queries first, then progressively exposed to more abstract formulations, we leave this exploration to future works.
\paragraph{Impact of Task Complexity}
\vspace{-2pt}
Figure~\ref{fig:steps_score} plots Soft Score against task complexity (the number of artifact transformations required). Performance declines with increasing complexity: single-artifact tasks are handled reasonably by most models, but tasks requiring 5+ transformations see scores collapse. This gap reveals that current frontier models struggle with long-horizon, multi-step spreadsheet workflows—a class of tasks that, while less frequent in the \wtmcorpus{} distribution (most tasks involve 1--2 transformations), represents precisely the high-value scenarios where automation would be most impactful.
% \paragraph{Artifact-Type Breakdown}
% \vspace{-2pt}
% Table~\ref{tab:artifact_breakdown} breaks down Soft Scores by artifact type. 
% Performance follows a clear hierarchy: \textbf{Formulas} are handled best (65.0\%), 
% followed by \textbf{Charts} (61.0\%), \textbf{Conditional Formatting} (36.0\%), 
% and \textbf{Pivot Tables} (10.0\%). The sharp drop at pivot tables reflects their 
% multi-step complexity---identifying source data, selecting fields, and choosing 
% aggregation functions. 

\paragraph{Artifact-Type Performance Breakdown}
\label{app:artifact_breakdown}

Table~\ref{tab:artifact_breakdown} presents the Soft Score breakdown by primary artifact type across models using VBA. Formulas and charts are handled competitively by most models, while conditional formatting and pivot tables remain challenging. Pivot tables in particular yield near-zero scores across all models, reflecting the multi-step coordination required (source data identification, field selection, aggregation specification, and output placement).

\begin{table}[h]
    \centering
    \setlength{\tabcolsep}{8pt}
    \renewcommand{\arraystretch}{1.15}
    \begin{tabular}{lcccc}
        \toprule
                \textbf{Model}& \textbf{Charts} & \textbf{Cond. Format.} & \textbf{Formulas} & \textbf{Pivot Tables} \\
        \midrule
        ClaudeHaiku45   & 51.0 & 18.0 & 43.0 & 4.0 \\
        ClaudeSonnet45  & 50.0 & 15.0 & 59.0 & \underline{9.0} \\
        ClaudeOpus46    & \underline{57.0} & \textbf{36.0} & \textbf{65.0} & \textbf{10.0} \\
        Gpt41Mini       & 52.0 & 2.0  & 32.0 & 0.0 \\
        Gpt52Reasoning  & 55.0 & \underline{33.0} & 59.0 & 5.0 \\
        Gpt54Reasoning  & \textbf{61.0}& 26.0 & \underline{62.0}& 8.0 \\
        \bottomrule
    \end{tabular}
    \caption{Soft Score (\%) results broken down by task artifact type.}
    \label{tab:artifact_breakdown}
\end{table}

\paragraph{Failure modes.} A rollout-level analysis over 4{,}910 deduplicated rollouts
(Appendix~\ref{app:robustness}) shows that failures (Soft Score -- failure < 0.5; strict failure < 0.10) are overwhelmingly structural rather
than numerical: among 1{,}858 failed Level~3 Python rollouts, 88.5\% either omit the
artifact entirely (62.3\%, wrong shape) or place it at the wrong coordinate (26.2\%,
wrong placement), while only 11.6\% produce a correctly shaped artifact with wrong values.
Tool-call failure is not the bottleneck, with frontier models siting at 3--5\%, and its
correlation with score is weak ($\rho = -0.11$). Code volume is the strongest negative
predictor ($\rho = -0.21$), suggesting that long generations often mark unsuccessful
repair attempts. Most consequentially for evaluation design, models frequently claim
completion despite strict failure: hallucinated success reaches 80.7\% for Claude
rollouts and 43.2\% for GPT.

% Notably, \wtmbench{} deliberately oversamples pivot tables 
% (15.3\% vs.\ 0.8\% in \wtmcorpus{}) to expose this gap, which would be invisible 
% in ecologically distributed evaluations.

\fi  % <<< end toggle
%%%%%%%%%%%%%%%%%%%%%%%%%%%%%%%%%%%%%%%%%%%%%%%%%%%%%%%%%%%%

\section{Conclusion}
We present the \emph{workbook time machine}, a pipeline for generating spreadsheet automation benchmarks from real-world Excel workbooks. Applied to public corpora, it produces \wtmcorpus{} (8{,}931 queries over 2{,}977 tasks), from which we curate \wtmbench{}, a balanced 150-task evaluation benchmark with near-uniform distribution across major Excel artifact types. Evaluation across 18 model-API combinations and 4 sheet-based orchestration reveals significant performance variations, demonstrating how API choice fundamentally impacts automation success, while our multi-level query design shows that frontier models achieve higher completion rates but face substantial challenges with ambiguous instructions and complex multi-step workflows, establishing robust evaluation standards that provide actionable insights for spreadsheet automation system improvement. 

\paragraph{Future Work}
The multi-level query design enables curriculum learning approaches, training models progressively from detailed Level 1 instructions to concise Level 3 queries to improve robustness to ambiguous instructions. The pipeline can be extended beyond creation tasks to support comprehensive CRUD operations, including deletion (by reversing input/output) and modification operations. Advanced workbook preprocessing techniques could further minimize structural information leakage through intelligent content repositioning and formula descriptor detection. Finally, the language-agnostic pipeline design allows straightforward adaptation to multilingual contexts, broadening applicability for international enterprise environments.

\newpage
\bibliographystyle{colm2026_conference}
\bibliography{references}

\newpage
\appendix

\section{Limitations and Scope}

The current version of \wtmbench{} focuses on derived analytical artifacts: formulas, charts, conditional formatting, and pivot tables. This scope is taxonomic rather than purely frequency-based. These artifacts transform source data into new analytical output, whereas data validation, filtering, sorting, tables, freezing panes, and related operations primarily control views, input constraints, or worksheet interaction state. Table~\ref{tab:feature_scope} reports the prevalence audit that informed this decision. The pipeline can be extended to these interaction-oriented operations, but they require different extraction and grading logic.

\begin{table}[h]
    \centering
    \small
    \setlength{\tabcolsep}{6pt}
    \begin{tabular}{lccc}
        \toprule
        \textbf{Feature} & \textbf{FUSE \%} & \textbf{Enron \%} & \textbf{In \wtmbench{}} \\
        \midrule
        Formulas & 4.60 & 58.09 & Yes \\
        Charts & 0.53 & 10.62 & Yes \\
        Conditional formatting & 0.45 & 2.98 & Yes \\
        Pivot tables & 0.039 & 1.58 & Yes \\
        Filtering & 0.83 & 3.91 & No \\
        Data validation & 0.75 & 0.54 & No \\
        Sorting & 0.71 & 0.00 & No \\
        Tables & 0.35 & 0.00 & No \\
        Sparklines & 0.00 & 0.00 & No \\
        \bottomrule
    \end{tabular}
    \caption{Prevalence of candidate Excel features in the source corpora. Sparklines were not observed in the audited FUSE workbooks.}
    \label{tab:feature_scope}
\end{table}

\wtmbench{} currently handles derivable objects with explicit formulas, while other derived content (such as manually entered lookup values) represents a distinct class of spreadsheet operations that could be addressed through extended formula descriptor analysis.

\section{Query Quality Evaluation}
\label{sec:quality}

\begin{figure*}[htbp]
    \centering
    \begin{subfigure}[b]{0.69\textwidth}
        \centering
        \includegraphics[width=\linewidth]{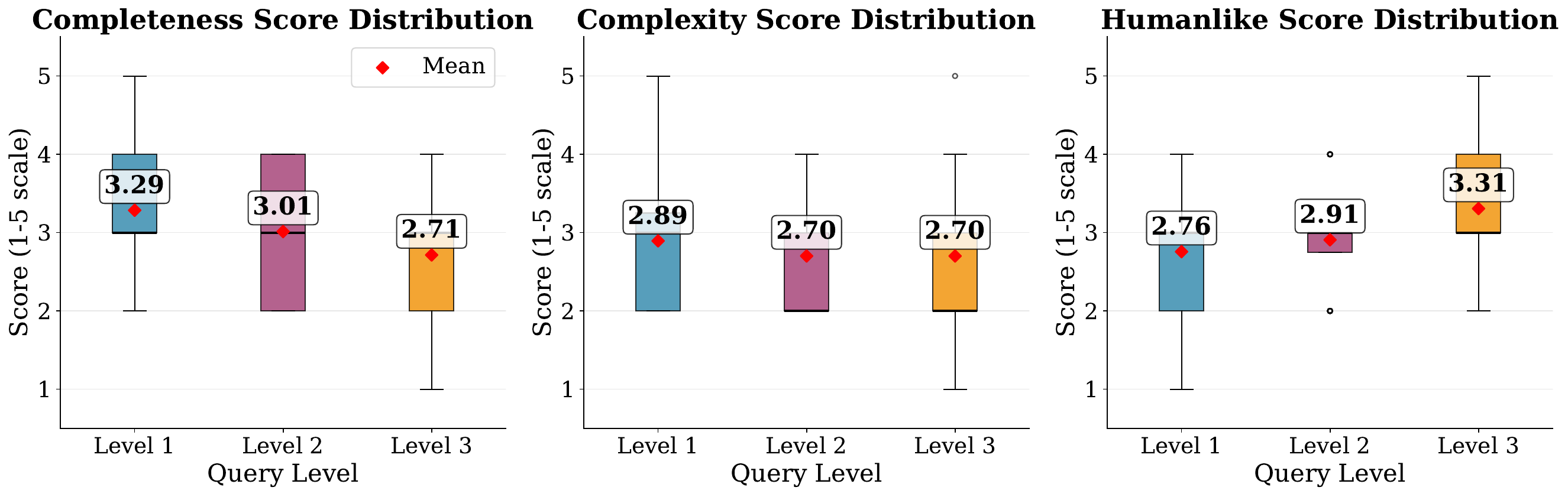}
        \caption{Quality score distributions across query levels}
        \label{fig:query_quality_distributions}
    \end{subfigure}
    \hfill
    \begin{subfigure}[b]{0.30\textwidth}
        \centering
        \includegraphics[width=\linewidth]{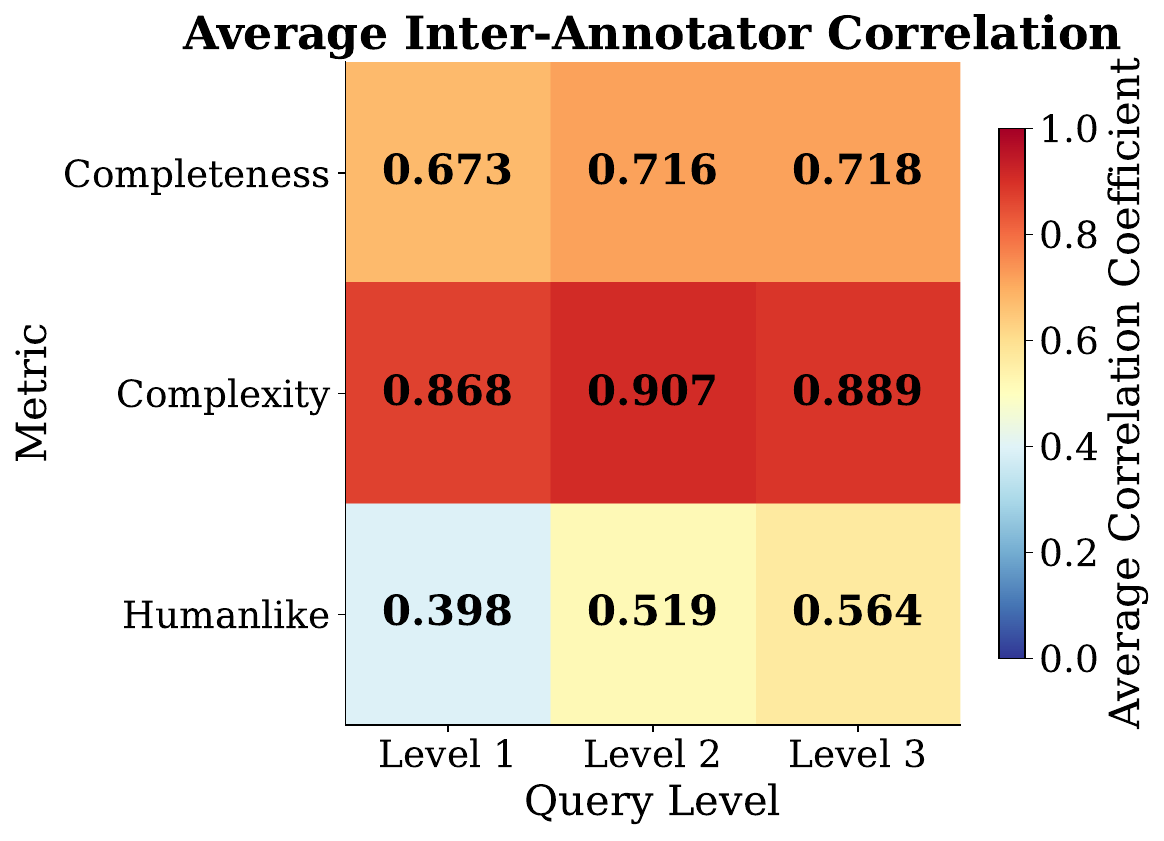}
        \caption{Inter-annotator agreement}
        \label{fig:correlation_heatmap}
    \end{subfigure}
    \caption{Query quality assessment: score distributions and annotation reliability analysis.}
    \label{fig:quality_analysis}
\end{figure*}

\paragraph{Multi-annotator quality assessment.}
We conducted a systematic quality evaluation using 3 human annotators, all paper authors with ML research backgrounds and spreadsheet-domain experience, across 75 sampled queries. Annotators independently rated completeness, complexity, and human-like quality on a 1--5 scale. Figure~\ref{fig:query_quality_distributions} presents score distributions across levels: completeness decreases with query level (3.29 to 2.71), complexity remains stable (2.89 to 2.70), while human-like scores increase (2.76 to 3.31). We interpret this as an indicative brevity--completeness trade-off, not as a correctness criterion for the benchmark.

\paragraph{Inter-annotator reliability.}
To ensure annotation quality, we assessed inter-annotator agreement across the 3 quality dimensions (Figure~\ref{fig:correlation_heatmap}). Complexity shows the highest agreement (average correlation: 0.888), indicating it is the most objectively assessable. Completeness shows moderate agreement (0.702), while human-like assessment has lower agreement (0.494), particularly for Level 1 queries (0.398), suggesting this dimension is more subjective. Completeness was scored as whether the query explicitly communicates details a human reader would consider necessary to execute the task, not whether the query uniquely determines the output workbook. Annotators may reasonably differ on whether axis titles, header names, or placements are essential or inferable from workbook context; therefore, this study characterizes query variants rather than validating task correctness. Correctness is instead determined by the programmatic grader in Appendix~\ref{app:metrics}.

\begin{figure}[htbp]
    \centering
    \includegraphics[width=\linewidth]{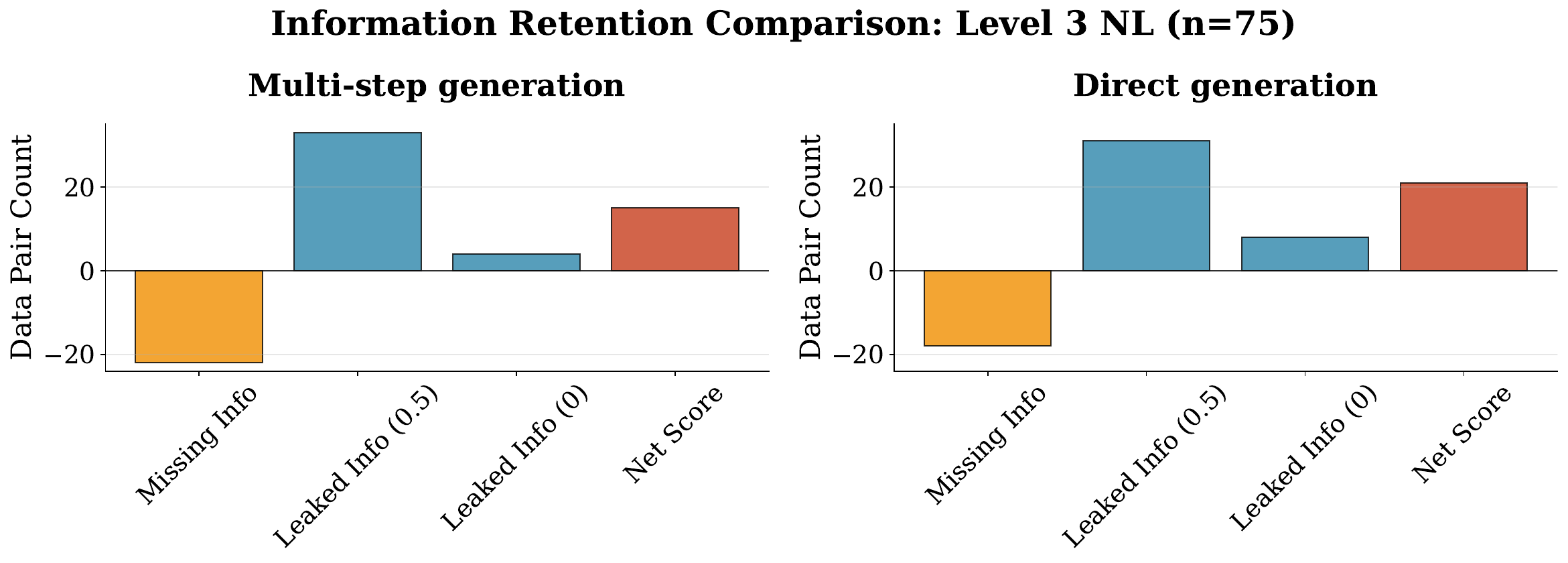}
    \caption{Information retention comparison between sequential multi-level generation (left) and direct Level 3 generation (right). Multi-level generation shows superior performance on essential information with fewer data leaks across 75 test cases.}
    \label{fig:retention_comparison}
\end{figure}

\section{Query Generation Pipeline: Design Decision Analysis}
\label{app:design_validation}

\paragraph{Motivation for multi-level generation.}
Given that Level 3 queries are the most under-specified, complex, and human-like in our taxonomy, our initial hypothesis was that generating all three levels (1 to 2 to 3) sequentially would provide better information retention and query quality than directly generating Level 3 queries. The rationale was that starting with verbose, well-specified Level 1 queries would preserve essential task details that could be progressively condensed while maintaining semantic fidelity.

\paragraph{Empirical comparison methodology.}
To validate this design decision, we conducted a systematic comparison of two generation approaches using 75 randomly sampled workbook tasks: (1) sequential generation of all three levels starting from detailed Level 1 queries and (2) direct generation of Level 3 queries without intermediate steps. We evaluated information retention using data-pair weighted metrics, tracking missing essential information (necessity=1.0) and leaked contextual details (necessity=0.5 and 0.0).

\paragraph{Findings.}
Figure~\ref{fig:retention_comparison} presents the information retention analysis results. The multi-level approach performs better than direct Level 3 generation across the key retention metrics. Direct generation has slightly more data pairs with missing critical information and a higher net retention score because it leaks more non-essential information. Ideally, each score should be close to 0; the multi-level method is closer to that target and yields better query quality.

We ran a small annotation over 75 samples to compare human-like scores for both these query generations: multi-level (3.44) vs. single (3.14), which clearly shows how multi-level generation is more aligned.

\begin{figure*}[htbp]
    \centering
    \begin{subfigure}[b]{0.69\textwidth}
        \centering
        \includegraphics[width=\linewidth]{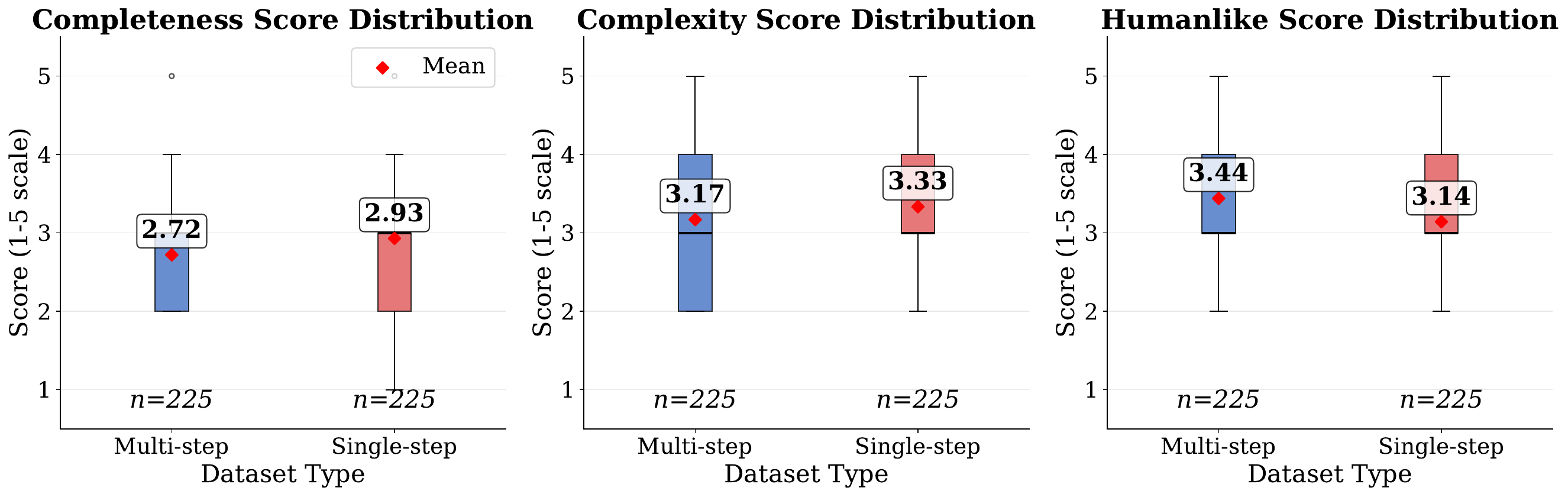}
        \caption{Quality score distributions across query levels}
        \label{fig:query_quality_distributions3}
    \end{subfigure}
    \hfill
    \begin{subfigure}[b]{0.30\textwidth}
        \centering
        \includegraphics[width=\linewidth]{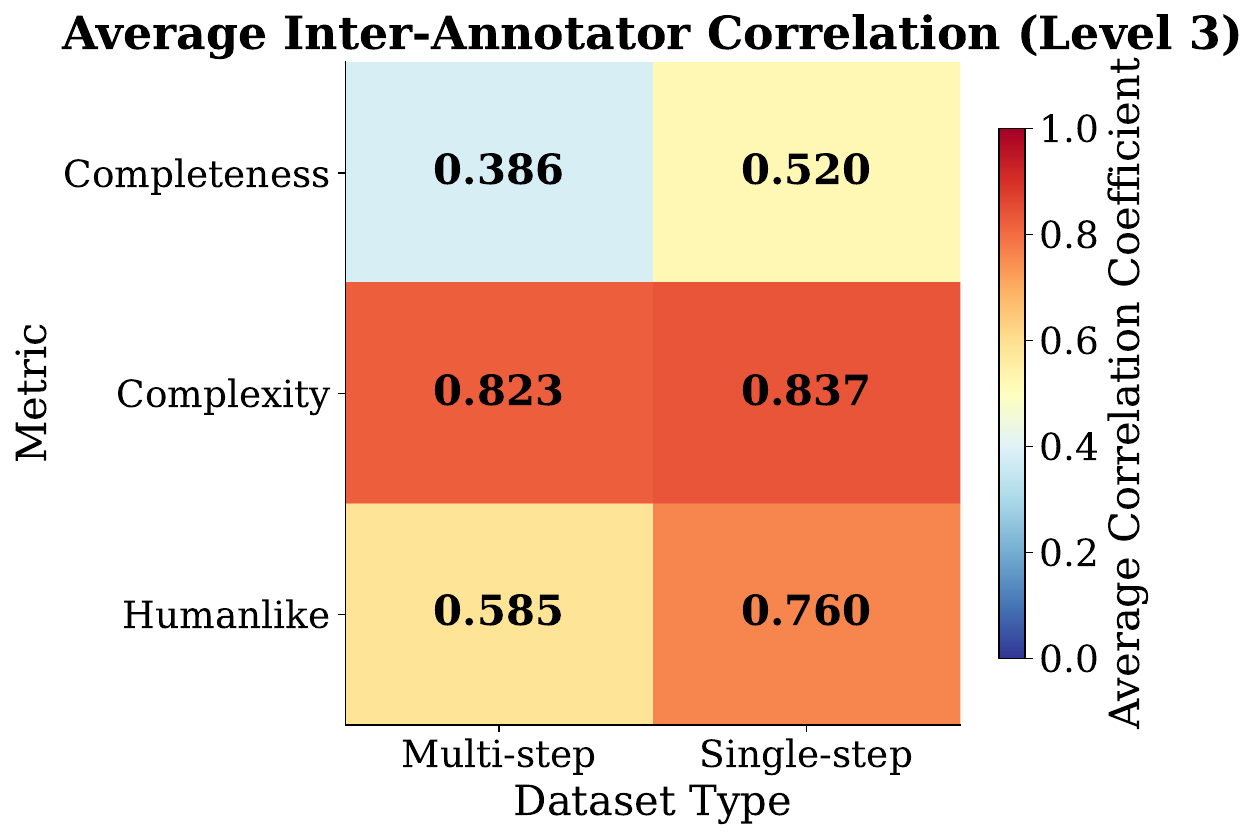}
        \caption{Inter-annotator agreement}
        \label{fig:correlation_heatmap3}
    \end{subfigure}
    \caption{Comparing query quality assessment over a multi-level vs. single-generation pass of 75 queries.}
    \label{fig:quality_analysis_multivs_single}
\end{figure*}

\section{Pipeline Construction Efficiency}
\label{app:construction_efficiency}

\begin{figure*}[htbp]
    \centering
    \begin{subfigure}[b]{0.49\textwidth}
        \centering
        \includegraphics[width=\linewidth]{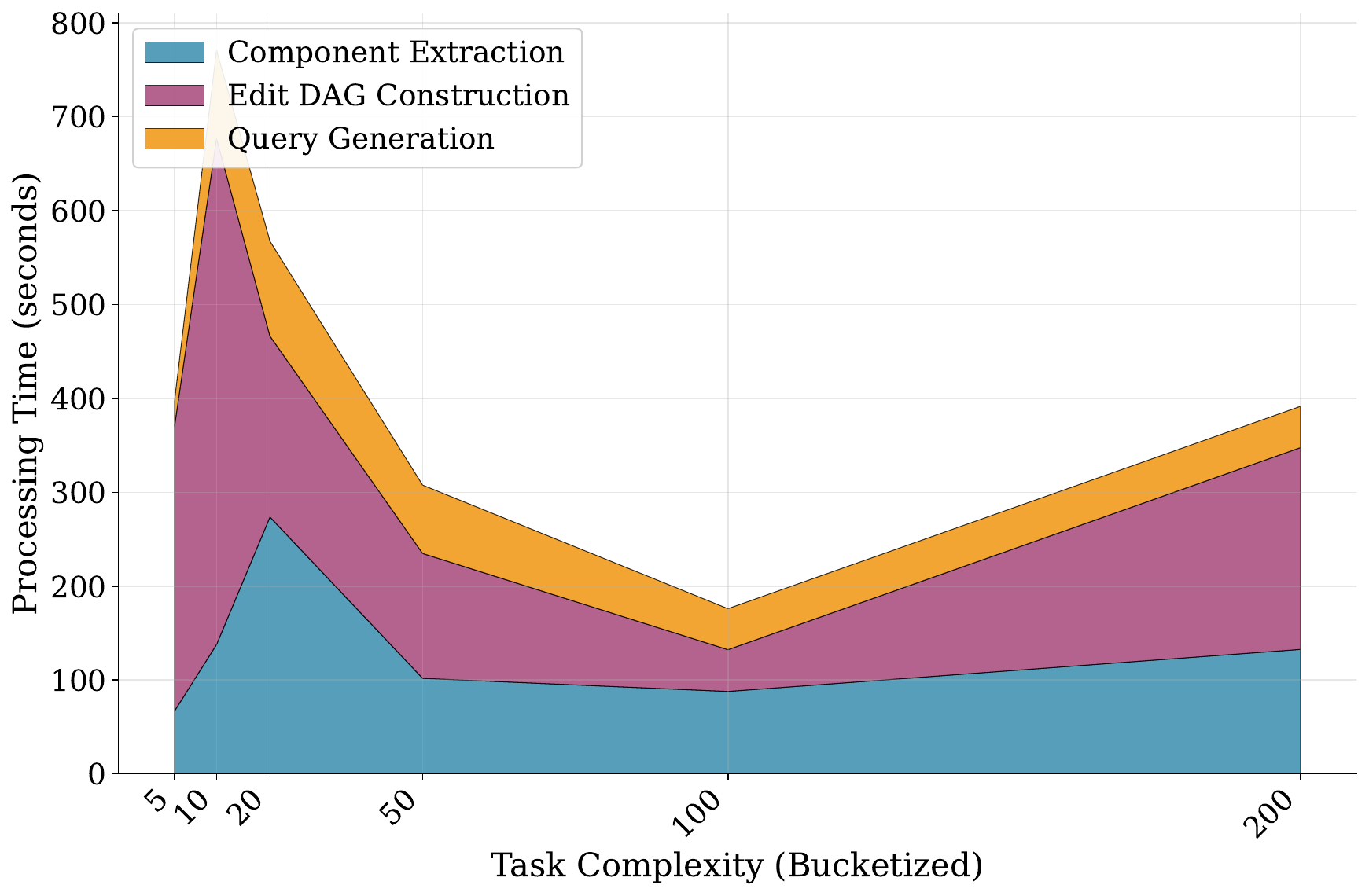}
        \caption{Computational time distribution across 3 pipeline stages and \# artifacts per workbook}
        \label{fig:pipeline_timing}
    \end{subfigure}
    \hfill
    \begin{subfigure}[b]{0.49\textwidth}
        \centering
        \includegraphics[width=\linewidth]{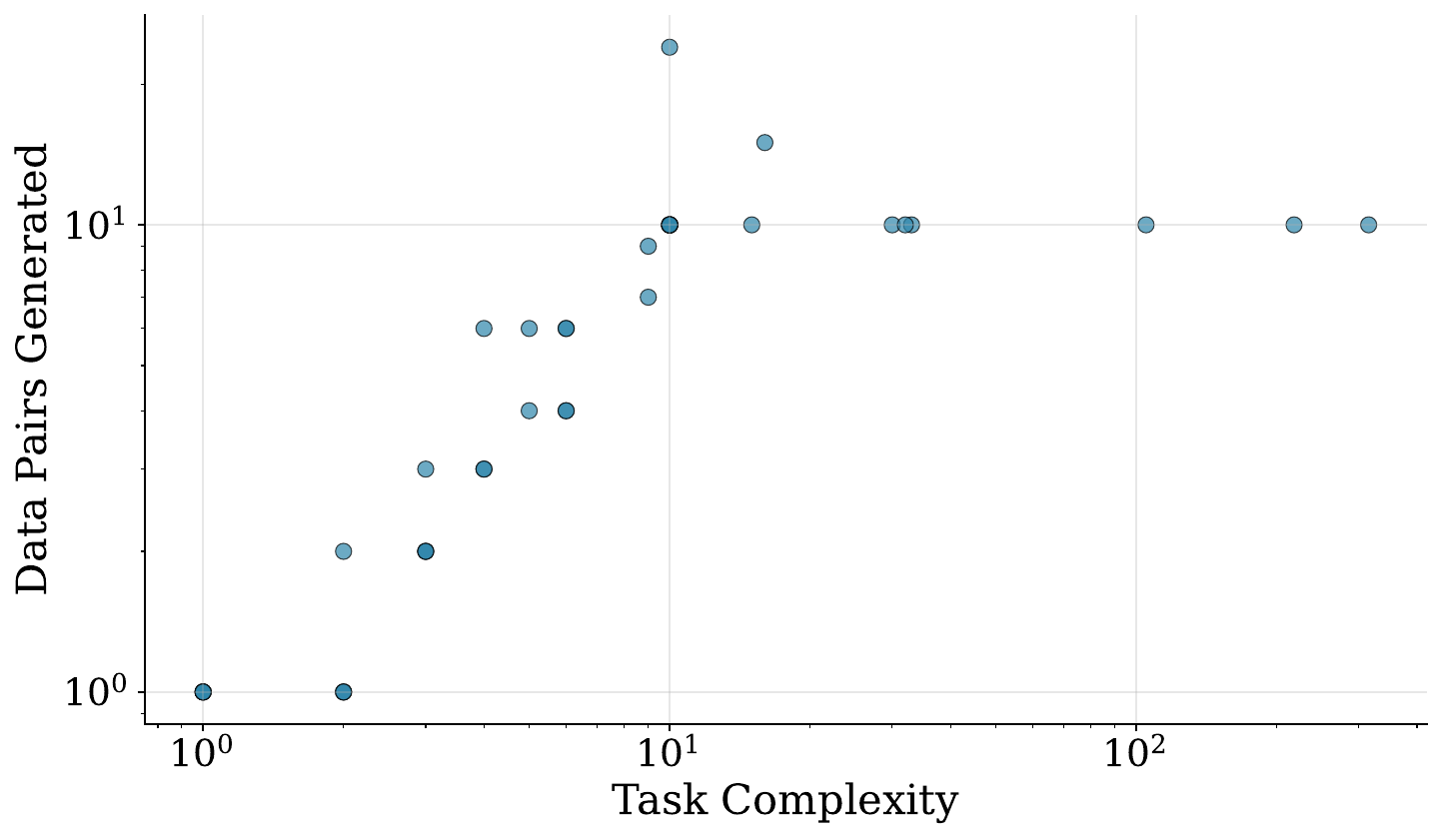}
        \caption{Logarithmic distribution of \# artifacts against the generated number of data pairs}
        \label{fig:time_traj}
    \end{subfigure}
    \caption{Pipeline construction efficiency: time distribution across stages (left) and scaling relationship between number of artifacts in workbook and generated data pairs (right).}
    \label{fig:time_scalability}
\end{figure*}

\paragraph{Construction time analysis.}
The mean processing time per workbook is 401\,s (median 192\,s) across our sample of 40 workbooks. Runtime is dominated by the Edit DAG construction stage (mean 254\,s, 63\% of total wall time), which enumerates and materializes all intermediate workbook states; Component extraction accounts for 25\% (mean 102\,s) and Query generation for 11\% (mean 44\,s). Figure~\ref{fig:time_scalability} demonstrates how total runtime scales with transformation complexity and data pair generation, confirming that richer workbooks amortize annotation cost across more training examples.

\paragraph{LLM usage scaling.}
For data generation, we use GPT-5.4-Reasoning due to its instruction-following capabilities, maintaining consistency throughout the pipeline. LLM usage scales directly with workbook complexity through the number of artifacts and worksheets requiring annotation. The decomposition $\text{LLM}_{\text{extract}} = \text{\# worksheets}$ for table identification and $\text{LLM}_{\text{annotate}} = 2 \times \text{\# data pairs} = 2 \times \sum \text{\# permutations of steps per trajectory}$ ensures that workbooks with richer artifact structures generate proportionally more training data while maintaining consistent annotation coverage.

\section{\wtmcorpus{} Distribution}\label{app:data_dist}

The workbook time machine generates benchmark instances from a large pool of real-world Excel workbooks, producing 8{,}931 natural-language queries over 2{,}977 unique tasks from the Enron and Fuse corpora. The full dataset pairs each task with multi-query variants at Levels~1, 2, and~3, enabling systematic evaluation across a verbosity-to-specificity spectrum while holding the underlying spreadsheet transformation fixed.

\subsection{Task-Type and Capability Distribution}

\begin{figure*}[htbp]
    \centering
    \begin{subfigure}[b]{0.29\textwidth}
        \centering
        \includegraphics[width=\linewidth]{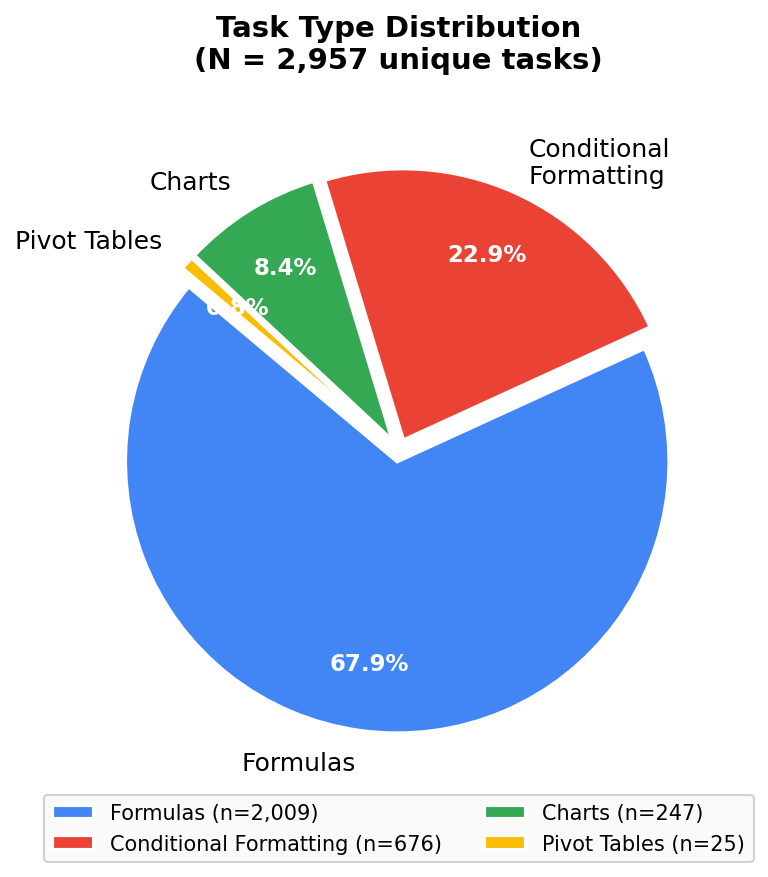}
        \caption{Task-type distribution ($N{=}2{,}957$).
                 Formulas account for 67.5\% of the full dataset.}
        \label{fig:task_type_pie}
    \end{subfigure}
    \hfill
    \begin{subfigure}[b]{0.70\textwidth}
        \centering
        \includegraphics[width=\linewidth]{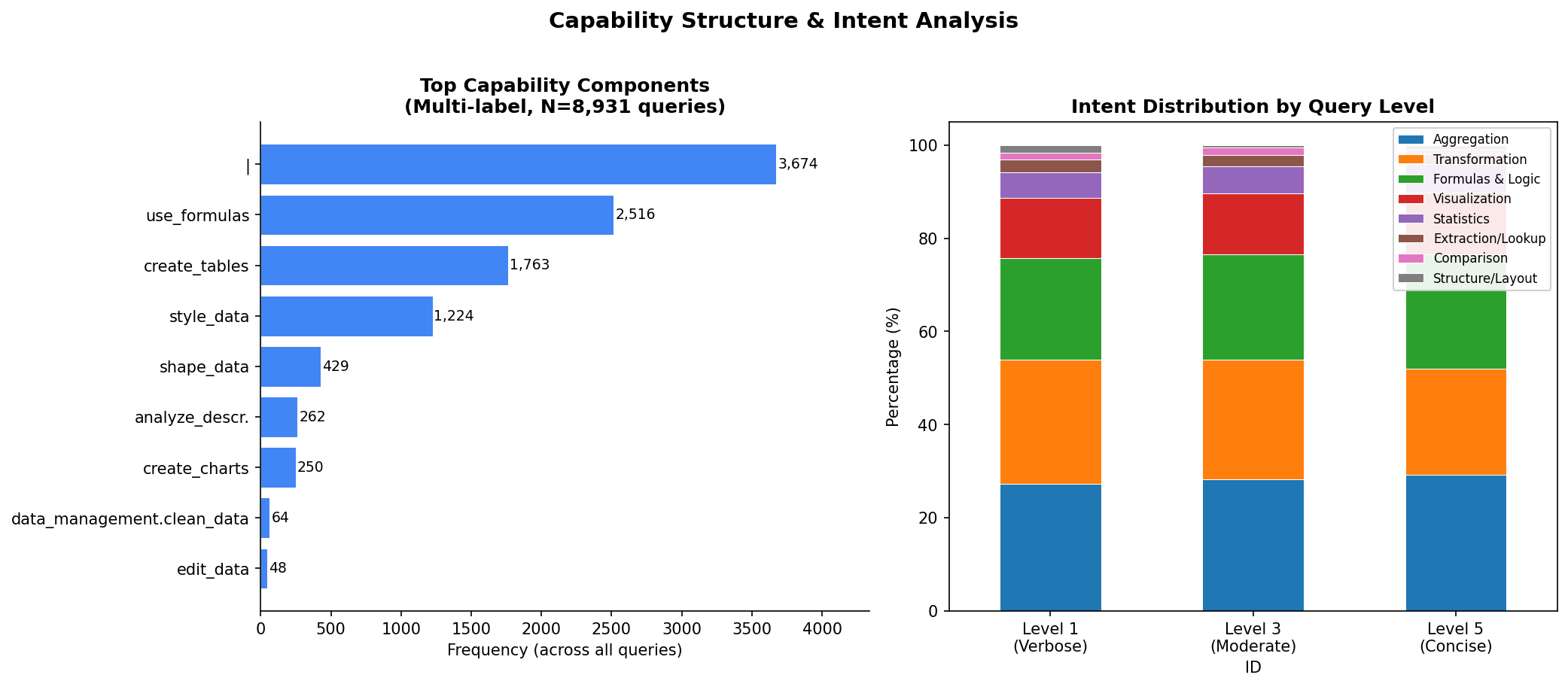}
        \caption{Artifact capability (left) and intent distribution
                 by query level (right).
                 Intent proportions are stable across Level~1--3,
                 confirming semantic equivalence of query variants.}
        \label{fig:capability_intent}
    \end{subfigure}
    \caption{Task-type and capability analysis of all query levels in the dataset (8{,}931 queries).}
    \label{fig:type_capability}
\end{figure*}

\paragraph{Task type composition.}
Figure~\ref{fig:task_type_pie} shows the distribution of the four primary task types across the 2{,}957 classified tasks (20 tasks could not be unambiguously categorized and are excluded from type-level analyses). Formulas constitute the largest category at 67.5\%, followed by Conditional Formatting (22.7\%), Charts (8.3\%), and Pivot Tables (0.8\%). This skewed distribution reflects the ecological reality of enterprise spreadsheet work, where formula-writing is the dominant activity and chart or pivot-table construction are specialist tasks performed comparatively rarely.

\paragraph{Capability structure.}
Each query is annotated with a set of \emph{capability} labels drawn from a structured ontology covering computation, data management, visualization, and analysis; most tasks require a combination (e.g.\ \texttt{use\_formulas} $+$ \texttt{style\_data}). Following established intent taxonomy methodologies~\citep{maruta2025spreadsheet, shah2025intent}, we categorize each query's functional intent. Figure~\ref{fig:capability_intent} (left) shows the frequency of individual artifact capability, and (right) the distribution of intent categories broken down by query level. The intent distribution is stable across levels: the proportional breakdown of Aggregation \& Summarization, Transformation, Formulas \& Logic, Visualization, and related intents remains consistent from Level~1 to Level~3, validating that the three query variants are semantically equivalent.

\subsection{Industry and Scenario Coverage}

\paragraph{NAICS sector coverage.}
The complete benchmark spans 20 distinct industry sectors coded under the North American Industry Classification System~\citep{naics2022}, providing cross-industry generalisation that prevents models from over-fitting to domain-specific vocabulary or conventions. The top five sectors--Utilities (33.2\%), Educational Services (19.7\%), Mining \& Oil/Gas (13.1\%), Finance \& Insurance (12.4\%), and Professional \& Technical Services (5.3\%)--account for approximately 84\% of tasks. The remaining 16\% spans 15 additional sectors including Healthcare, Manufacturing, Real Estate, Public Administration, and Transportation.

\paragraph{Financial workflow scenarios.}
Each task is labelled with a \emph{financial scenario} from a ten-category vocabulary that captures functional spreadsheet intent. The most frequent scenarios are Revenue Analysis (14.3\%), Financial Modelling (12.8\%), Treasury \& Cash Flow (7.3\%), Budgeting \& Forecasting (7.2\%), and Expense Tracking (4.5\%). Formulas are distributed nearly uniformly across all scenario types, consistent with their role as a general-purpose tool; Charts concentrate in Basic Analysis and Visualisation scenarios; and Pivot Tables concentrate in Advanced Analysis.

\begin{figure*}[htbp]
    \centering
    \includegraphics[width=\linewidth]{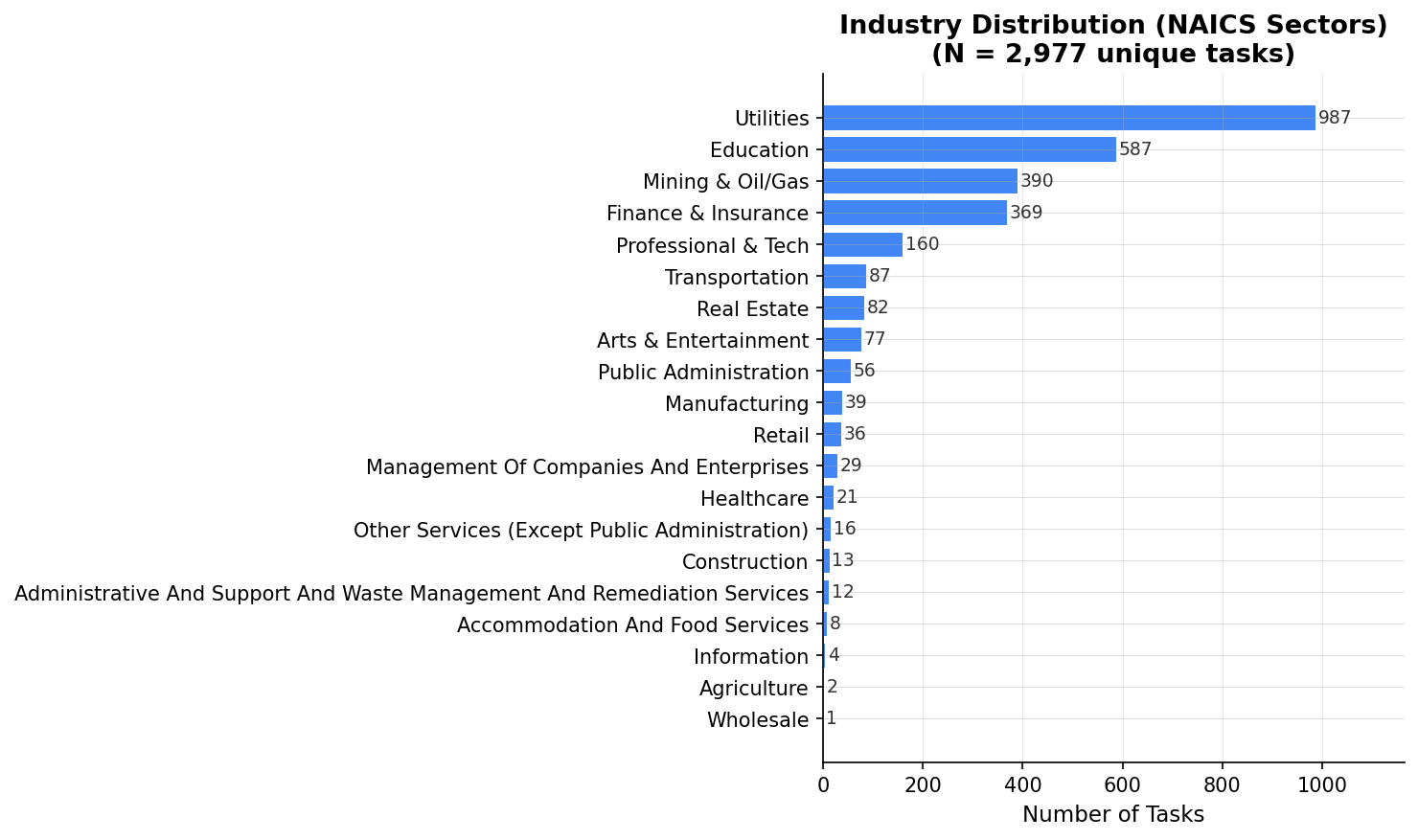}
    \caption{NAICS sector and financial scenario distributions across the complete 3k dataset.}
    \label{fig:industry_distribution_3k}
\end{figure*}

\subsection{Query-Level Specificity Gradient}

The complete dataset implements a three-level query design where each of the 2{,}977 unique tasks is described by three semantically equivalent natural language queries at distinct abstraction levels: Level~1 (verbose, mean 357 characters), Level~2 (moderate, mean 220 characters), and Level~3 (concise, mean 139 characters). The proportion of \emph{Very Well-Specified} queries drops from 77.9\% at Level~1 to 17.8\% at Level~3--a 4.4$\times$ reduction--while Ambiguous or Reasonably Specified queries increase from 0\% to 3.1\%. Query length contracts monotonically across levels, and Level~1 queries imply more steps on average (mean $= 3.96$) than Level~3 (mean $= 2.99$) despite requiring the same underlying action.

\begin{figure*}[htbp]
    \centering
    \includegraphics[width=\linewidth]{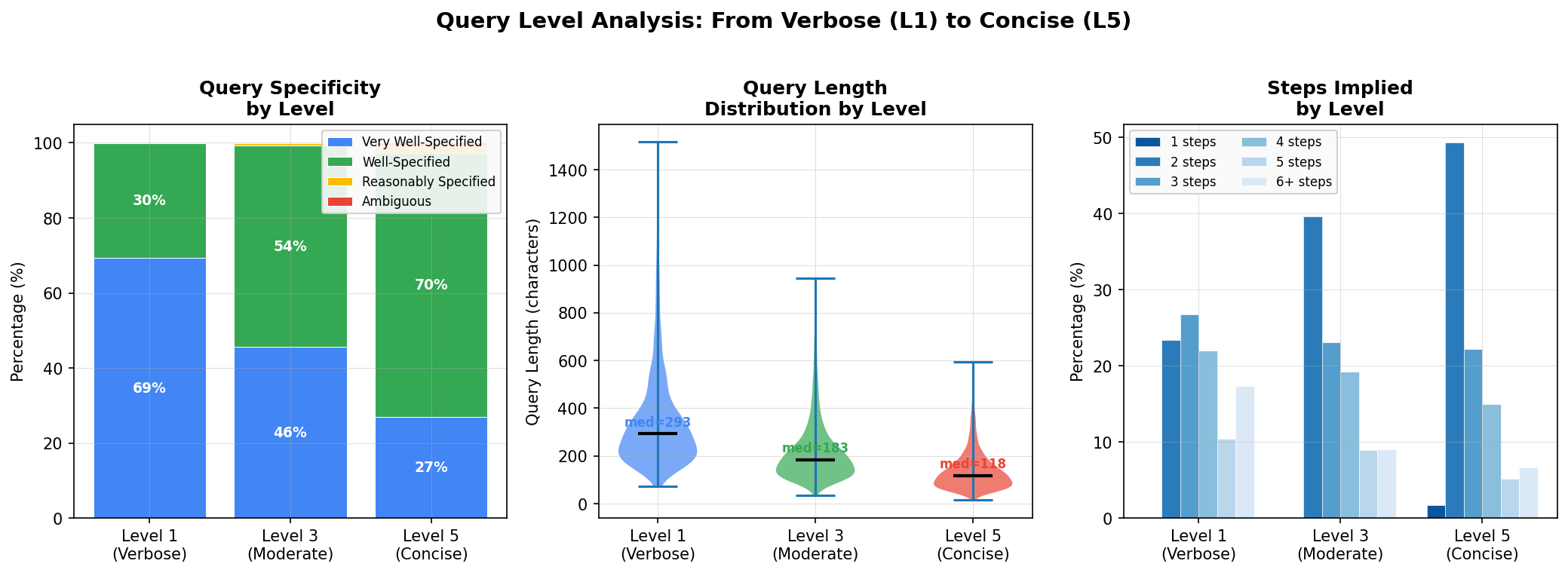}
    \caption{Level analysis across the complete dataset: specificity (left), query length (centre), implied steps (right) by query level.}
    \label{fig:level_analysis_3k}
\end{figure*}

\section{Detailed Experimental Specifications}
\label{app:exp_details}

\subsection{Model Specifications}

We evaluate 6 state-of-the-art language models from two major families:

\paragraph{Anthropic models:}
\textbf{Claude Haiku 4.5} is a fast, lightweight model optimized for simple tasks with efficient processing capabilities. \textbf{Claude Sonnet 4.5} represents a balanced performance model designed for general use cases, offering good trade-offs between capability and efficiency. \textbf{Claude Opus 4.6} serves as the most capable model in the family, specifically designed for complex reasoning tasks requiring sophisticated analytical capabilities.

\paragraph{OpenAI models:}~\citep{openai2023gpt4}
\textbf{GPT-4.1 Mini} provides an efficient model architecture tailored for straightforward tasks where computational efficiency is prioritized. \textbf{GPT-5.2 Reasoning} incorporates advanced reasoning capabilities designed to handle more complex logical operations and multi-step problem-solving. \textbf{GPT-5.4 Reasoning} represents the state-of-the-art reasoning model, offering the highest level of analytical and inferential capabilities in the evaluated model suite.

\subsection{API Details}

We evaluate 3 API interfaces for Excel manipulation:

\paragraph{Python} OpenPyXL~\citep{openpyxl} is a Python library providing direct programmatic access to Excel files. It offers comprehensive support for reading and writing Excel 2010 xlsx/xlsm files, including formulas, charts, and conditional formatting. However, limitations include lack of VBA macro support, missing Excel object primitives and reduced performance with very large files. We also provide the pandas library in the environment, to help with Pivot Tasks, which OpenPyXL cannot natively perform.

\paragraph{VBA} serves as Excel's native macro language with full feature access to all Excel functionality. It provides complete control over Excel objects, methods, and properties, enabling sophisticated automation scenarios. The primary advantages include comprehensive feature coverage and direct Excel integration without external dependencies. Limitations encompass Windows-only operation and a steeper learning curve compared to modern programming interfaces.

\paragraph{OfficeJS~\citep{officejs}} represents Microsoft's modern JavaScript API for web-based interaction with Office applications. Designed for cross-platform compatibility and modern web development practices, it enables spreadsheet automation through web-based interfaces. Advantages include robust cross-platform support and modern API design principles. However, limitations include reduced functionality compared to VBA and dependency on web-based execution environments.

\subsection{Orchestration Framework}\label{app:orches}

\paragraph{Multi-Turn Agent Loop.}
The core evaluation driver implements a multi-turn tool-calling loop. At each turn, the model receives the conversation history and may either produce a final text response (terminating the task rollout) or emit a structured tool call containing executable code. Each of these outputs can be accompanied with an optional <think>...</think> paragraph. Tool call outputs--including execution results and the sheet state (post code execution)--are appended to the conversation as tool-role messages, and the loop continues for up to a maximum number of 10 turns. 

\paragraph{Tool Execution Backends.}
The framework supports three interchangeable code execution backends, each defined by a distinct system prompt, function-calling schema, and executor.

\begin{itemize}
    \item \textbf{Python (OpenPyXl):} Code is executed in-process via \texttt{exec()} in a sandboxed daemon thread with a timeout.

    \item \textbf{OfficeJS:} Code is executed through an Excel runtime.

    \item \textbf{VBA:} Code is executed through COM automation (\texttt{win32com}).
\end{itemize}

Each backend returns a structured result containing an execution status flag and the post-execution sheet state.

\section{Robustness and Failure Analysis}
\label{app:robustness}

\paragraph{Independent re-runs.}
To test whether the main results are stable under API nondeterminism, we ran 3 additional independent trials on the most evaluated configuration, Python/OpenPyXL Level~3, across 7 models. Table~\ref{tab:rerun_variance} reports mean $\pm$ standard deviation. Model ordering is preserved across runs and Hard Score standard deviations are at most 0.035.

\begin{table}[h]
    \centering
    \small
    \setlength{\tabcolsep}{4pt}
    \begin{tabular}{lccc}
        \toprule
        \textbf{Model} & \textbf{\# Turns} & \textbf{Soft Score} & \textbf{Hard Score} \\
        \midrule
        Claude Haiku 4.5 & $5.50 \pm 0.05$ & $0.353 \pm 0.016$ & $0.103 \pm 0.012$ \\
        Claude Sonnet 4.5 & $4.95 \pm 0.09$ & $0.431 \pm 0.019$ & $0.157 \pm 0.009$ \\
        Claude Opus 4.6 & $4.31 \pm 0.02$ & $0.510 \pm 0.017$ & $0.180 \pm 0.010$ \\
        GPT-4.1 Mini & $2.18 \pm 0.17$ & $0.294 \pm 0.026$ & $0.083 \pm 0.016$ \\
        GPT-5.4 Mini & $1.66 \pm 0.04$ & $0.351 \pm 0.014$ & $0.111 \pm 0.010$ \\
        GPT-5.2 Reasoning & $3.58 \pm 0.12$ & $0.480 \pm 0.024$ & $0.160 \pm 0.012$ \\
        GPT-5.4 Reasoning & $2.66 \pm 0.45$ & $0.437 \pm 0.020$ & $0.148 \pm 0.035$ \\
        \bottomrule
    \end{tabular}
    \caption{Python/OpenPyXL Level~3 runs over 150 tasks per run with 0--1 scale.}
    \label{tab:rerun_variance}
\end{table}

\paragraph{Turn budget.}
The 10-turn budget is generally permissive rather than binding. Mean turns range from 1.66 to 5.50 in the repeated Python setting and from 2.31 to 4.61 for the reported OfficeJS/VBA runs. The two smaller Claude models hit the cap on a minority of hard Python tasks (Haiku: 24.2\%; Sonnet: 17.8\%), while GPT models rarely do so. Thus, the single-turn SpreadsheetBench gap is not primarily a turn-budget artifact.

\begin{table}[h]
    \centering
    \small
    \setlength{\tabcolsep}{4pt}
    \begin{tabular}{lccc}
        \toprule
        \textbf{Model} & \textbf{Python} & \textbf{OfficeJS} & \textbf{VBA} \\
        \midrule
        Claude Haiku 4.5 & 5.50 (24.2\%) & 4.61 (15.3\%) & 3.16 \\
        Claude Sonnet 4.5 & 4.95 (17.8\%) & 4.18 (14.0\%) & 2.92 \\
        Claude Opus 4.6 & 4.31 (7.1\%) & 4.08 (6.4\%) & 3.50 \\
        GPT-4.1 Mini & 2.18 (1.1\%) & 2.69 (4.7\%) & 2.20 \\
        GPT-5.4 Reasoning & 2.66 (0.4\%) & 2.63 (1.3\%) & 2.31 \\
        \bottomrule
    \end{tabular}
    \caption{Mean turns by API; parentheses show fraction of rollouts that reached 10-turn cap.}
    \label{tab:turn_budget}
\end{table}

\paragraph{Failure taxonomy.}
We analyzed 4{,}910 deduplicated rollouts covering Python Level~3 (150 tasks $\times$ 3 runs $\times$ 7 models), Python Level~1, and OfficeJS. Failure is defined as Soft Score below 0.5; strict failure is below 0.10. Among 1{,}858 failed Level~3 Python rollouts, 88.5\% are structural: either the artifact is effectively absent (wrong shape) or present at the wrong coordinate (wrong placement). Wrong values account for only 11.6\% (Table~\ref{tab:error_verticals}).
\begin{table}[h]
    \centering
    \small
    \begin{tabular}{lccc}
        \toprule
        \textbf{Error vertical} & \textbf{$n$} & \textbf{Share} & \textbf{95\% CI} \\
        \midrule
        Wrong shape & 1{,}157 & 62.3\% & [60.1, 64.3] \\
        Wrong placement & 486 & 26.2\% & [24.3, 28.1] \\
        Wrong value & 215 & 11.6\% & [10.2, 13.0] \\
        \bottomrule
    \end{tabular}
    \caption{Grader-derived error verticals for failed Level~3 Python rollouts.}
    \label{tab:error_verticals}
\end{table}
Both model families have similar error-vertical splits; their main difference is hallucinated success, where the final message claims completion despite strict failure. As Table~\ref{tab:family_failure} shows, Claude rollouts have higher hallucinated success and more turns on average, while GPT rollouts terminate earlier.
\begin{table}[h]
    \centering
    \small
    \begin{tabular}{lcccc}
        \toprule
        \textbf{Family} & \textbf{Soft mean} & \textbf{Hard mean} & \textbf{Hallucinated success} & \textbf{Mean turns} \\
        \midrule
        Claude & 43.9 $\pm$ 0.5\% & 14.5 $\pm$ 0.3\% & 80.7\% & 4.92 \\
        GPT & 39.4 $\pm$ 0.6\% & 12.3 $\pm$ 0.8\% & 43.2\% & 2.44 \\
        \bottomrule
    \end{tabular}
    \caption{Family-level failure behavior in the rollout analysis.}
    \label{tab:family_failure}
\end{table}
Execution-surface correlations (Table~\ref{tab:surface_correlations}) show that tool failure is not the main bottleneck: frontier models sit at 3--5\% tool-call failure, and score correlation with tool-call failure is weak. Code volume is the strongest negative predictor, suggesting that long code generations often mark unsuccessful repair attempts. Exploration polarity differs by family: read-only exploration correlates negatively with score for Claude and positively for GPT.
\begin{table}[h]
    \centering
    \small
    \setlength{\tabcolsep}{4pt}
    \begin{tabular}{lrrrr}
        \toprule
        \textbf{Surface} & \textbf{Overall} & \textbf{Claude} & \textbf{GPT} & \textbf{Small tier} \\
        \midrule
        Tool-call failure & $-0.114$ & $-0.118$ & $-0.114$ & $-0.238$ \\
        Code volume & $-0.213$ & $-0.327$ & $-0.186$ & $-0.272$ \\
        Number of turns & $-0.076$ & $-0.299$ & $-0.006$ & $-0.172$ \\
        Exploration ratio & $0.021$ & $-0.216$ & $0.106$ & $-0.038$ \\
        Effective recovery & $-0.111$ & $-0.154$ & $-0.083$ & $-0.205$ \\
        \bottomrule
    \end{tabular}
    \caption{Spearman correlations between execution surfaces and score.}
    \label{tab:surface_correlations}
\end{table}
Across query levels, Soft Score drops modestly from Level~1 to Level~3, but Hard Score drops more sharply (for example, Opus falls from 27\% to 18\%, and GPT-5.4 Reasoning from 21\% to 13\%). Thus Level~3 primarily reduces fully correct rollouts rather than only lowering partial credit. Across APIs, OfficeJS improves Soft Score slightly for all models, but GPT tool-call failure increases by 17 percentage points while Claude remains stable, exposing an execution weakness that aggregate Soft Score alone masks.

\section{Detailed Grading Schema}
\label{app:metrics}

All components compare only changes from the pre-task baseline (initial $\rightarrow$ ground truth vs.\ initial $\rightarrow$ generation); pre-existing content is ignored. Each component is active only when the ground truth introduces that artifact type, and raw weights are normalized to sum to 1.0. Supporting signals (worksheet creation and tables) carry half the weight of primary signals.

\paragraph{Cell Values \& Formulas:}
Cell values are evaluated by computing formula results (via the Formulas library), so equivalent expressions that produce the same output receive full credit. 

\paragraph{Charts:}
Each ground-truth chart is matched to the best available generated chart. Chart similarity score is computed across three criteria: chart type (40\%: full credit for exact match, 20\% for same family, e.g.\ \texttt{BarChart} vs.\ \texttt{BarChart3D}), series count (20\%: proportional partial credit via min/max ratio), and series data references (40\%: Jaccard overlap of reference strings). Axis labels, legend styling, and color choices are not evaluated.

\paragraph{Pivot Tables:}
Pivot similarity is scored across five sub-criteria: sheet placement (15\%), row fields (25\%: full credit for correct ordered list, 15\% for correct fields in wrong order), column fields (20\%: full credit for correct ordered list, 10\% for correct fields in wrong order), data field names (25\%: Jaccard set overlap), and cache fields / source schema (15\%: Jaccard overlap of source column names). Aggregation function names are not directly evaluated; the data field name match serves as a proxy.

\paragraph{Conditional Formatting:}
Scoring covers three criteria: correct worksheet (30\%), exact cell range after normalisation (40\%), and rule type set overlap (30\%, e.g.\ \texttt{cellIs}, \texttt{colorScale}, \texttt{dataBar}). Formatting style properties such as colors, fonts, and borders are not evaluated, nor is rule priority ordering.

\begin{figure}
    \begin{subfigure}[b]{0.52\linewidth}
        \centering
        \includegraphics[width=\linewidth]{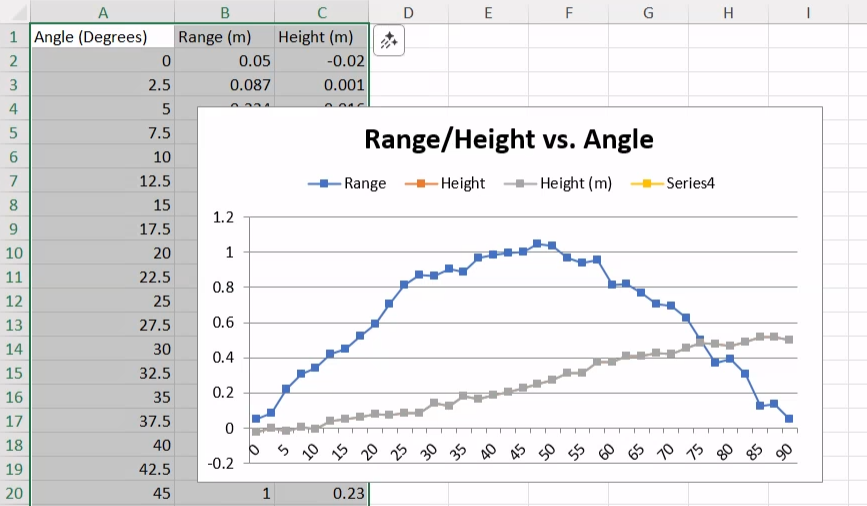}
        \caption{Generated}
    \end{subfigure}
    \hfill
    \begin{subfigure}[b]{0.42\linewidth}
        \centering
        \includegraphics[width=\linewidth]{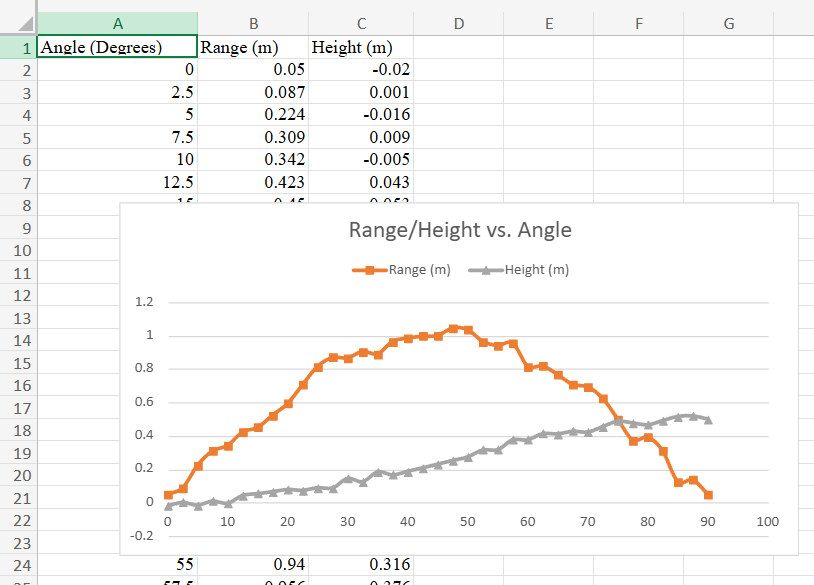}
        \caption{Ground Truth}
    \end{subfigure}
    \caption{Evaluation metrics comparison. For it, Hard match: \textcolor{red}{Fail}, Soft match: \textcolor{green}{Pass}.}
    \label{fig:evalexample}
\end{figure}

\section{Examples from \wtmbench{}}
\begin{figure}
    \centering
    \includegraphics[width=\linewidth]{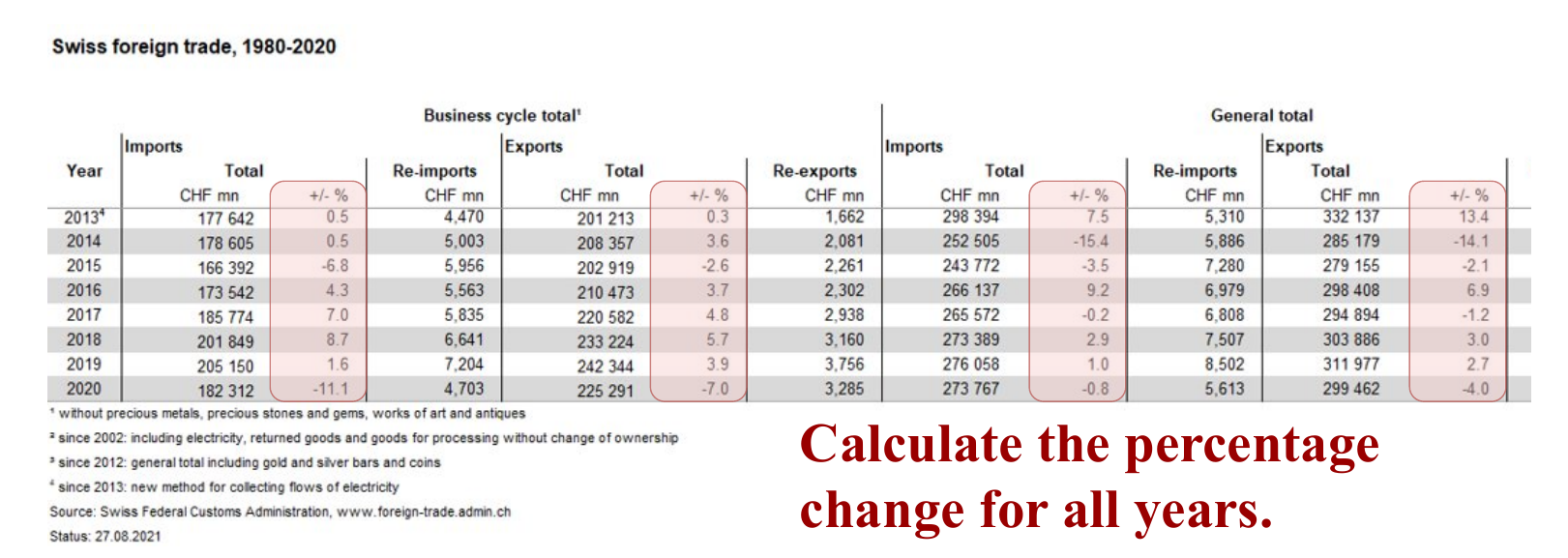}
    \caption{Example 1: Clubbed formula group being added. Where the colored texts are the utterances corresponding to the task.}
    \label{fig:example1}
\end{figure}

\begin{figure*}
    \centering
    \begin{subfigure}{0.52\linewidth}
        \centering
        \includegraphics[width=\linewidth]{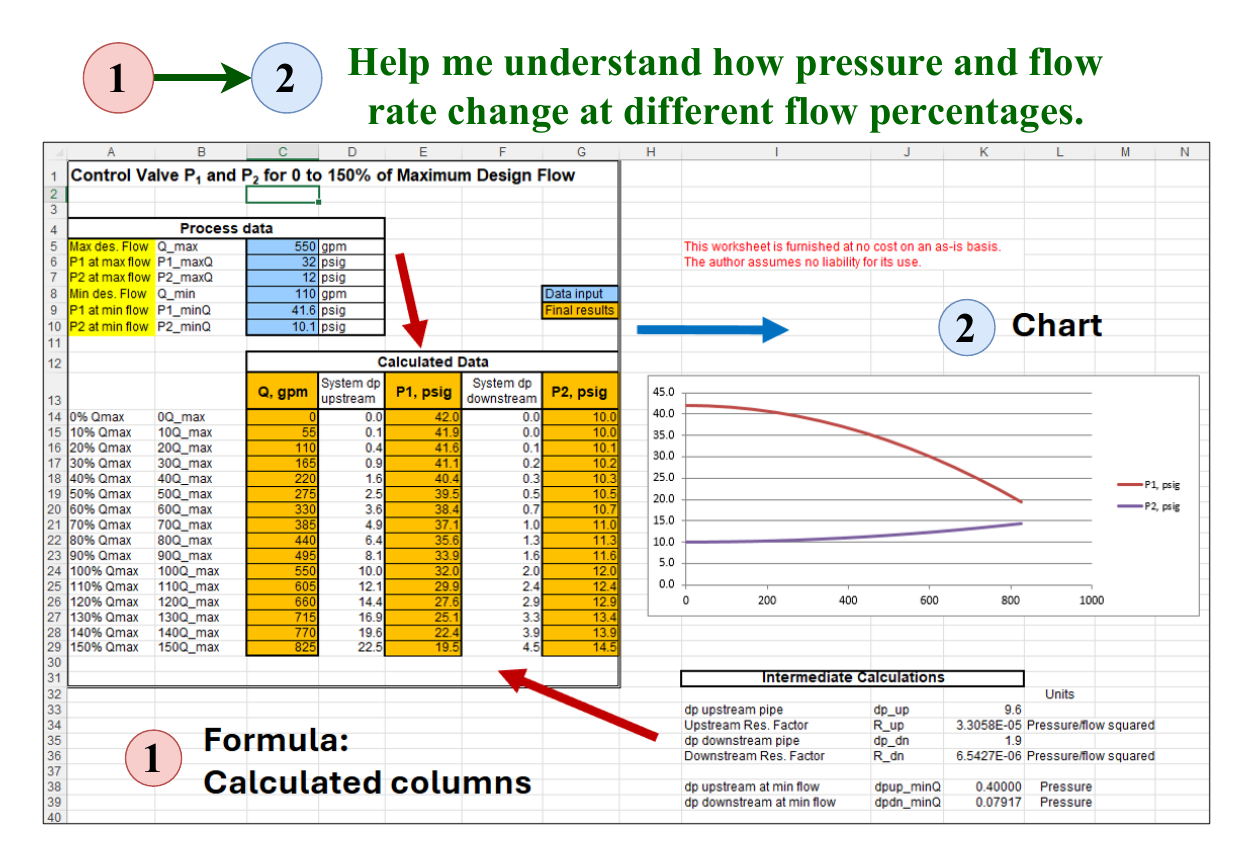}
        \caption{Mixed multi-step tasks: formulas, charts}
        \label{fig:example2}
    \end{subfigure}
    \hfill
    \begin{subfigure}{0.46\linewidth}
        \centering
        \includegraphics[width=\linewidth]{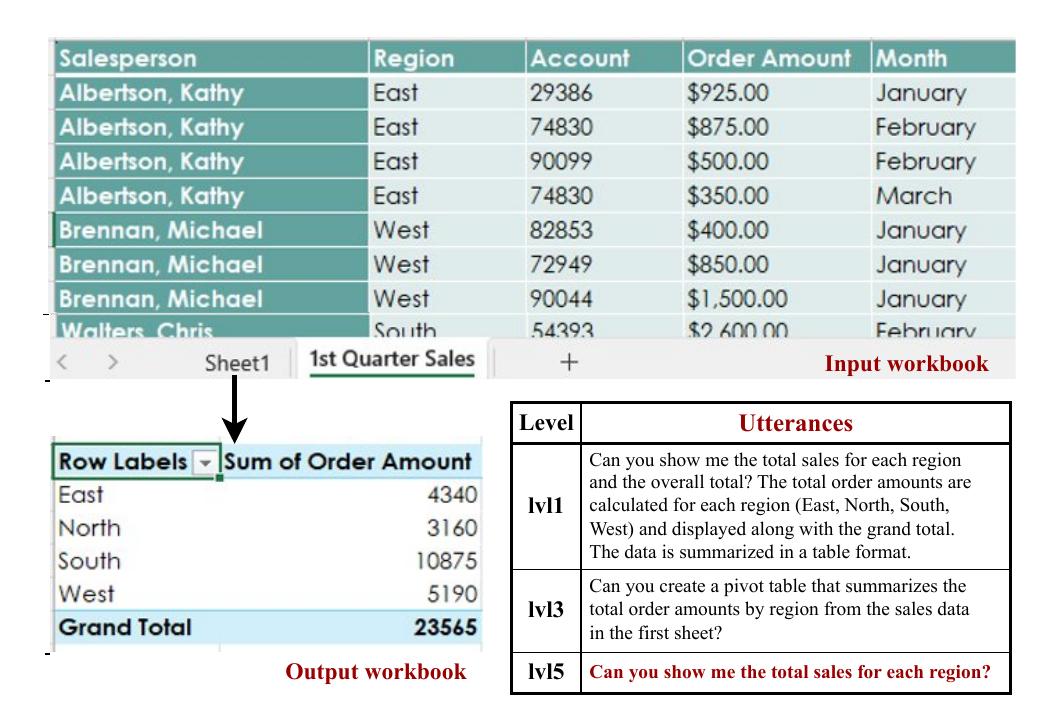}
        \caption{Pivot table on new worksheet}
        \label{fig:example3}
    \end{subfigure}
    \caption{Examples of complex spreadsheet tasks}
    \label{fig:examples_combined}
\end{figure*}

\begin{figure*}
    \centering
    \includegraphics[width=0.72\linewidth]{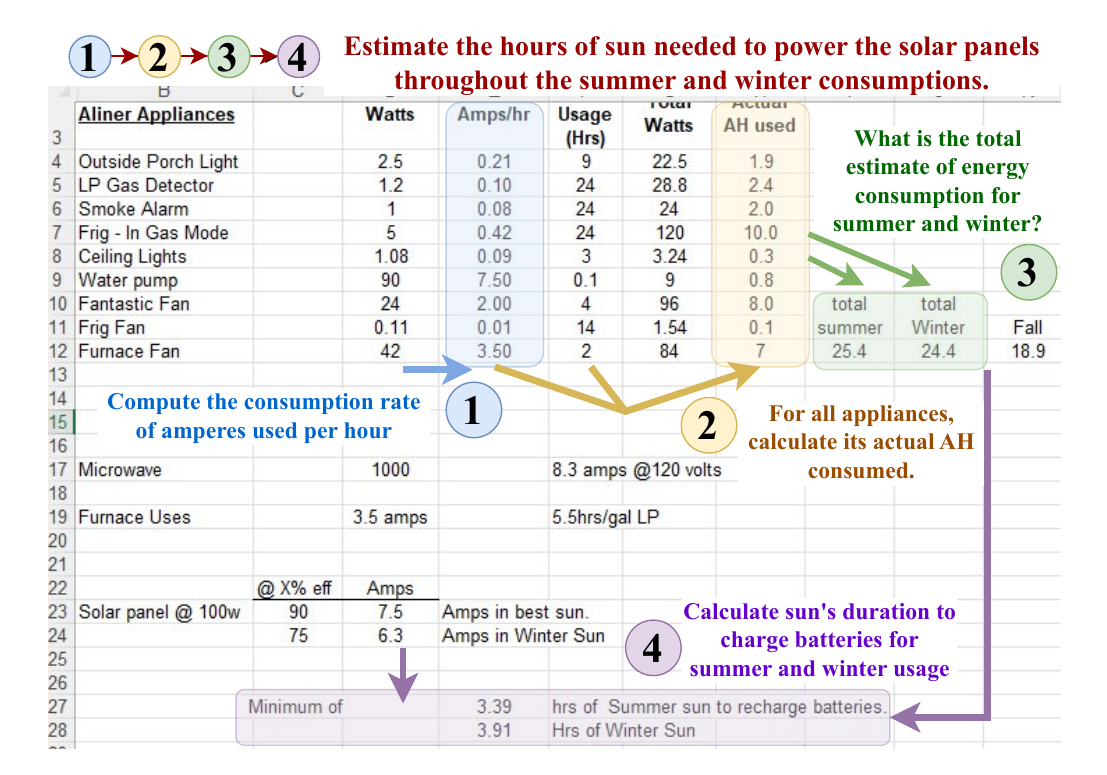}
    \caption{Example 4: Multi-step formula tasks}
    \label{fig:example4}
\end{figure*}

Examples from \wtmbench{} are presented in the subsequent Figs. \ref{fig:example1}, \ref{fig:examples_combined}, and \ref{fig:example4}. 

\end{document}